\pdfoutput=1

\documentclass[letterpaper]{article} 
\usepackage{aaai23}  
\usepackage{times}  
\usepackage{helvet}  
\usepackage{courier}  
\usepackage[hyphens]{url}  
\usepackage{graphicx} 
\usepackage{natbib}  
\usepackage{caption} 
\usepackage{algorithm}
\usepackage{algorithmic}

\usepackage{enumitem}
\usepackage{amsmath,amssymb} 
\usepackage{multirow}
\usepackage{booktabs}
\usepackage{subcaption}
\usepackage{xspace}
\usepackage{xcolor}
\usepackage{xspace}
\usepackage{enumerate}
\usepackage{enumitem}
\usepackage{amsfonts}

\usepackage[title]{appendix}

\usepackage{newfloat}
\usepackage{listings}
\DeclareCaptionStyle{ruled}{labelfont=normalfont,labelsep=colon,strut=off} 
\floatstyle{ruled}
\newfloat{listing}{tb}{lst}{}
\floatname{listing}{Listing}
\title{Dynamic Ensemble of Low-fidelity Experts: Mitigating NAS ``Cold-Start''}
\author{
  Junbo Zhao\equalcontrib,\textsuperscript{\rm 1,\rm 3}
    Xuefei Ning\equalcontrib\thanks{Corresponding authors.},\textsuperscript{\rm 1}
    Enshu Liu,\textsuperscript{\rm 1}
    Binxin Ru,\textsuperscript{\rm 4}
    Zixuan Zhou,\textsuperscript{\rm 1}\\
    Tianchen Zhao,\textsuperscript{\rm 1}
    Chen Chen,\textsuperscript{\rm 2}
    Jiajin Zhang,\textsuperscript{\rm 2}
    Qingmin Liao,\textsuperscript{\rm 3}
    Yu Wang$^\dagger$\textsuperscript{\rm 1}
}
\affiliations{
    \textsuperscript{\rm 1}Department of Electronic Engineering, Tsinghua University\\
    \textsuperscript{\rm 2}Huawei Technologies Co., Ltd\\
    \textsuperscript{\rm 3}Tsinghua Shenzhen International Graduate School\\
    \textsuperscript{\rm 4}SailYond Technology  \& Research Institute of Tsinghua University in Shenzhen\\
    $^\dagger$foxdoraame@gmail.com, $^\dagger$yu-wang@tsinghua.edu.cn
}

\usepackage{bibentry}

\begin{document}

\maketitle

\begin{abstract}
Predictor-based Neural Architecture Search (NAS) employs an architecture performance predictor to improve the sample efficiency. However, predictor-based NAS suffers from the severe ``cold-start'' problem, since a large amount of architecture-performance data is required to get a working predictor. In this paper, we focus on exploiting information in cheaper-to-obtain performance estimations (i.e., low-fidelity information) to mitigate the large data requirements of predictor training. Despite the intuitiveness of this idea, we observe that using inappropriate low-fidelity information even damages the prediction ability and different search spaces have different preferences for low-fidelity information types. To solve the problem and better fuse beneficial information provided by different types of low-fidelity information, we propose a novel dynamic ensemble predictor framework that comprises two steps. In the first step, we train different sub-predictors on different types of available low-fidelity information to extract beneficial knowledge as low-fidelity experts. In the second step, we learn a gating network to dynamically output a set of weighting coefficients conditioned on each input neural architecture, which will be used to combine the predictions of different low-fidelity experts in a weighted sum. The overall predictor is optimized on a small set of actual architecture-performance data to fuse the knowledge from different low-fidelity experts to make the final prediction. We conduct extensive experiments across five search spaces with different architecture encoders under various experimental settings. For example, our methods can improve the Kendall's Tau correlation coefficient between actual performance and predicted scores from 0.2549 to 0.7064 with only 25 actual architecture-performance data on NDS-ResNet. Our method can easily be incorporated into existing predictor-based NAS frameworks to discover better architectures. Our method will be implemented in Mindspore~\citep{mindspore}, and the example code is published at \url{https://github.com/A-LinCui/DELE}.
\end{abstract}

\section{Introduction}
In recent years, architectures automatically designed by neural architecture search (NAS)~\cite{elsken2019neural} have achieved state-of-the-art performance on various tasks~\cite{zoph2016neural,liu2018darts,chen2019detnas,wang2020fcos}. Accurate and efficient architecture performance estimation strategy is one of the key components of NAS~\cite{elsken2019neural}, which can be broadly divided into 
parameter-sharing-based~\cite{pham2018efficient} and predictor-based methods~\cite{luo2018neural,ning2020generic,white2021powerful}. The former evaluates with weights shared in an over-parametrized super network, while the latter learns a predictor to predict the performance of candidate architectures.

Predictor-based NAS trains an approximate performance predictor and utilizes it to rank unseen architectures without actually training them. Therefore, once we have a predictor that can reliably rank the performance of unseen architectures, the architecture exploration can be significantly accelerated. However, predictor-based NAS suffers from the severe ``cold-start'' problem: It usually takes quite a considerable cost to acquire the architecture-performance data needed for training a working predictor from scratch. 

Recognizing the high cost of getting actual architecture-performance data as the major challenge for predictor-based NAS, existing efforts seek to learn the predictor in a more data-efficient way. Researchers have designed specialized predictor architectures~\cite{ning2020generic,zhang2019dvae,tang2020semi,yan2021cate,ning2022tagates}, training losses~\cite{luo2018neural,ning2020generic,xu2021renas,tang2020semi,yan2020does,yan2021cate}, to exploit information in the limited data more efficiently. In contrast, \textbf{our work focuses on exploiting more information in other cheaper-to-obtain performance estimations (i.e., low-fidelity information) to mitigate the data requirements of predictor training.} Actually, it is intuitive that utilizing other low-fidelity information (e.g., grasp~\cite{wang2020picking} and plain~\cite{1988Skeletonization}) for predictor training can help mitigate the cold-start problem. One can anticipate that training with this information might bring potential improvements in two aspects. On the one hand, the ranking information included in some indicators (e.g., {one-shot~\cite{pham2018efficient}, zero-shot~\cite{abdelfattah2021zerocost,Lin2021zen} estimations}) might help the predictor acquire a better ranking quality. On the other hand, learning to fit other low-fidelity information could encourage the predictor to extract better architecture representations. 

\begin{table*}[t]
\centering
\resizebox{\linewidth}{!}{
\begin{tabular}{c|ccccc}
\toprule
\multirow{2}{*}{\textbf{Low-fidelity Type}} & \multicolumn{5}{c}{\textbf{Low-fidelity Corr. /  Kendall's Tau Relative Improvement}}\\
\cmidrule(lr){2-6} & \textbf{NAS-Bench-201}&\textbf{NAS-Bench-301} &\textbf{NDS-ResNet}&\textbf{NDS-ResNeXt-A} &\textbf{MobileNet-V3}\\
\midrule
\midrule
\textbf{grasp}~\cite{wang2020picking} & +0.3227 / -0.0118  & +0.4062 /  +0.0189 & -0.1142 / -0.9431 & -0.2615 / -1.0230 & -0.0663 / +0.0091 \\
\textbf{plain}~\cite{1988Skeletonization} & -0.1467 / -0.1423  & -0.4670 / -0.7677 & +0.3066 / +0.7477 & +0.2887 / +0.4064 & +0.0116 / -0.0544 \\
\textbf{synflow}~\cite{tanaka2020pruning} & +0.5808 / +0.1463 & +0.1967 / -0.2653 & +0.2307 / +0.7270 & +0.6904 / +1.0406 & +0.6366 / -0.0024 \\
\textbf{grad\_norm}~\cite{abdelfattah2021zero} & +0.4798 / +0.1597 & +0.0378 / -0.3340 & +0.2372 / +0.8286 & +0.3190 / +0.6247 & +0.0696 / +0.0238 \\
\textbf{jacob\_cov}~\cite{Mellor2020NASwithoutTraining} & +0.4763 / +0.1780 & +0.0958 / -0.1654 & -0.0724 / -0.5551 & +0.0510 / -0.5681 & -0.0053 / -0.1423 \\
\midrule
\bottomrule
\end{tabular}
}
\caption{The ``Low-Fidelity Corr.'' and relative Kendall's Tau improvement achieved by utilizing different typical types of low-fidelity information.
Specifically, we construct the predictor with an LSTM encoder and train it with ranking loss. All architectures in the training split are used for pretraining, while the first 1\% percentages by index with corresponding actual performance are used for finetuning. ``Low-Fidelity Corr.'' represents Kendall's Tau correlation between the low-fidelity information and the actual performance.}
\label{tab:diff_ss_diff_lf}
\end{table*}

A straightforward way of utilizing low-fidelity information is to pretrain the model on a single type of low-fidelity information and finetune it on a small amount of actual architecture-performance data. We conduct a preliminary experiment in Table~\ref{tab:diff_ss_diff_lf} and make the following observations.

\begin{itemize}[leftmargin=*]
\item \emph{Low-fidelity information does have the potential to improve prediction ability with limited actual architecture-performance data significantly. E.g., utilizing grad\_norm increases the relative Kendall's Tau\footnote{Kendall's Tau is the relative difference of the number of concordant pairs and discordant pairs, reflecting the ranking correlation between predictions and ground-truths.} for 0.1597 and 0.8286 on NAS-Bench-201~\cite{dong2020bench} and NDS-ResNet~\cite{radosavovic2019nds}, respectively.}
\item \emph{Inappropriate low-fidelity information types even damage the prediction ability. E.g., utilizing ``plain'' decreases the relative Kendall's Tau for 0.1423, 0.7677 on NAS-Bench-201 and NAS-Bench-301~\cite{siems2020bench}, respectively.}
\item \emph{Different search spaces have different preferences for low-fidelity information types. E.g., grad\_norm~\cite{abdelfattah2021zero} decreases Kendall's Tau on NAS-Bench-301 but benefits the prediction on the other search spaces.}
\item \emph{A high-ranking quality of the low-fidelity information does not indicate its utilization effectiveness. E.g., synflow~\cite{tanaka2020pruning} positively correlates with actual performance but damages prediction on NAS-Bench-301.}
\end{itemize}

That is to say, despite the intuitiveness of this idea, which types of low-fidelity information are useful for performance prediction is unclear to practitioners beforehand. In addition, different types of low-fidelity information could provide beneficial information from different aspects, but the naive method described above can only utilize one type of low-fidelity information. Therefore, it would be better if we could fuse the knowledge from multiple types of low-fidelity information organically in an automated way.

In this paper, \textbf{we propose a novel dynamic ensemble predictor framework}, whose core is a learnable gating network that maps the neural architecture to a set of weighting coefficients to be used in ensembling predictions of different low-fidelity experts. The framework comprises two steps. In the first step, we pretrain different low-fidelity experts on different types of available low-fidelity information to extract beneficial knowledge. In the second step, the overall predictor is finetuned on the actual architecture-performance data to fuse knowledge from different types of low-fidelity information to make the final prediction. In this way, we can not only leverage multiple low-fidelity information in the architecture performance prediction but also balance their contributions in an automatic and dynamic fashion, overcoming the challenge for the practitioners to decide on which low-fidelity information to use.

To demonstrate the effectiveness of our proposed method, we conduct extensive experiments across multiple benchmarks, including NAS-Bench-201, NAS-Bench-301, NDS ResNet, NDS ResNeXt-A~\cite{radosavovic2019nds}, and MobileNetV3~\cite{cai2019once}. And our experiments are conducted under various experimental settings (\emph{e.g.,} different predictor construction methods, varying data sizes). 
We show that our method of exploiting additional low-fidelity information can significantly and consistently improve the ranking quality of predictors compared to using the architecture-performance solely, thus improving the overall NAS efficiency. Our method can be easily incorporated into existing predictor-based NAS methods to alleviate the cold-start problem and guide the NAS process to discover better architectures. For example, our dynamic ensemble predictor discovers architectures with 94.37\% test accuracy on NAS-Bench-201 (CIFAR-10~\cite{krizhevsky2009learning}), surpassing ReNAS~\cite{xu2021renas} (93.99\%) and NEPNAS~\cite{wei2020npenas} (91.52\%) with the same search budget. More information is available on our website \url{https://sites.google.com/view/nas-nicsefc/home/search-strategy-improvement/dele}.

\section{Related Work}
\label{sec:rw}
\subsection{Fast Evaluation Strategies in NAS}
Neural architecture search (NAS)~\cite{elsken2019neural} is a technique to design neural network architectures automatically. The vanilla NAS method~\cite{zoph2016neural} is computationally expensive since it needs to train each candidate architecture from scratch to get its performance. Therefore, a series of methods focus on developing faster architecture evaluation strategies to address the computational challenge. The two most popular types of fast evaluation strategies are the one-shot estimators~\cite{bender2018understanding,pham2018efficient,guo2019single} and zero-shot estimators~\cite{Mellor2020NASwithoutTraining,abdelfattah2021zerocost}.

\subsubsection{One-shot performance estimations.} 
One-shot NAS methods~\cite{bender2018understanding,pham2018efficient,guo2019single} construct an over-parametrized network (namely supernet), in which all candidate architectures are contained and share weights. After being trained to convergence, the supernet can evaluate the performance of each architecture by directly using the corresponding weights. Due to its efficiency, the one-shot performance estimation strategy is widely studied and used on different search spaces~\cite{cai2019once,wu2019fbnet} and for different tasks~\cite{chen2019detnas,wang2020fcos}. 
However, as reported in EEPE~\cite{ning2020surgery}, one-shot performance estimations might have unsatisfying correlation and prominent bias. Therefore, one-shot performance estimation can fail to 
benefit NAS \cite{pourchot2020share}. 

\subsubsection{Zero-shot performance estimations.} Recently, several researches~\cite{Mellor2020NASwithoutTraining,abdelfattah2021zerocost,Lin2021zen} propose ``zero-shot'' estimators, which utilize randomly initialized weights to estimate architectures' performance. Since no training process is required, these estimations are extremely fast. Nevertheless, EEPE~\cite{ning2020surgery} reveals that these zero-shot estimations have prominent biases, no zero-shot estimator can get a satisfying ranking quality in all search spaces, and the best zero-shot estimator is different across search spaces.

\subsection{Predictor-based NAS}
Predictor-based NAS~\cite{luo2018neural,ning2020generic,wei2020npenas,tang2020semi,xu2021renas,white2021powerful,ning2022tagates} is another type of NAS methods that relies on an architecture performance predictor. 
An architecture performance predictor takes the architecture description as the input and outputs an estimated score. In each iteration of predictor-based NAS, the predictor is trained on actual architecture-performance data and then utilized to efficiently evaluate and sample new architectures. Then, the architecture-performance data of the newly sampled architectures would be used to tune the predictor in the next iteration. 
The most costly part of the predictor-based NAS flow is getting the actual architecture-performance data for predictor training. 
We refer the readers to the GATES paper \cite{ning2020generic} for a summary of the general predictor-based NAS workflow. Recently, Wu \emph{etal.} \cite{wu2021stronger} derives a formulation for predictor-based NAS and justify the rationality of this widely-used workflow.

A problem with predictor-based NAS is that we usually need many actual architecture-performance data to get a working predictor. The initial exploration in the search space is poorly guided and usually just conducted by random sampling. We refer to this problem as the ``cold-start problem''. 

\subsubsection{Improving Predictor-based NAS.}
Researchers have been focused on making the predictor utilize available data more efficiently. Existing methods can be resolved into two aspects: 1) \textbf{The construction of predictor architectures}: NASBot~\cite{kandasamy2018neural} employs Gaussian Process as the predictor to better model the uncertainty. For topological search spaces, graph-based predictors are designed to encode the architecture in a better way~\cite{ning2020generic,dudziak2020brp,shi2020bridging,ning2022tagates}. Tang \emph{etal.} \cite{tang2020semi} propose explicitly modeling the relation between multiple architectures to predict their performances. 2) \textbf{The loss design of predictor training}: GATES 
\cite{ning2020generic} and ReNAS 
\cite{xu2021renas} propose to train the predictor with ranking loss to provide better architecture comparison. Several other studies~\cite{luo2018neural,tang2020semi,yan2020does} employ reconstruction loss as an auxiliary loss term. 

\subsubsection{Utilizing Cheaper-to-obtain Estimations.}
Recently, several studies have attempted to exploit cheaper-to-obtain performance estimations in predictor-based NAS to improve search efficiency: 1) ProxyBO \cite{shen2021proxybo} proposes to combine the architecture ranking given by the predictor and zero-shot proxies in the search process; 2) White \emph{etal.} \cite{white2021powerful} find that certain families of performance estimations can be combined to achieve even better predictive power; 3) AceNAS \cite{zhang2021acenas} proposes to pretrain the predictor with FLOPs, parameter size, and weight-sharing accuracy in a multi-task manner. However, these methods either heavily rely on the high-ranking correlation between the actual performance and the utilized estimations \cite{shen2021proxybo} or require carefully utilized estimation selection \cite{white2021powerful, zhang2021acenas}. Different from these attempts, our method has no requirement for correlation between the utilized estimations and actual performance and can use a broader range of cheaper-to-obtain estimations without the need for manual hand-picking.

\subsection{Low-Fidelity Information}
Low-fidelity information refers to indicators obtained with a low computational cost. These indicators capture some properties of neural architectures and thus can indicate their performances to some extent. We anticipate that learning to fit these cheaper-to-obtain data can encourage the predictor to extract better architecture representations and thus boost the ranking quality of the predictor. Different types of low-fidelity information can be roughly classified as follows.
\begin{itemize}[leftmargin=*]
\item \textbf{One-shot information.} The performance of architectures obtained from the one-shot supernet.
\item \textbf{Zero-shot information}, such as grad\_norm, synflow and synflow\_bn~\cite{tanaka2020pruning}, snip~\cite{Lee2018snip}, grasp, fisher~\cite{Theis2018faster} and jacob\_cov~\cite{Mellor2020NASwithoutTraining}.
\item \textbf{Complexity information.} Architecture information from the complexity perspective, such as the number of floating-point operations (FLOPs), the parameter size (params), and the inference latency (latency).
\end{itemize}


\begin{figure*}[t]
\center
\includegraphics[width=0.9\linewidth]{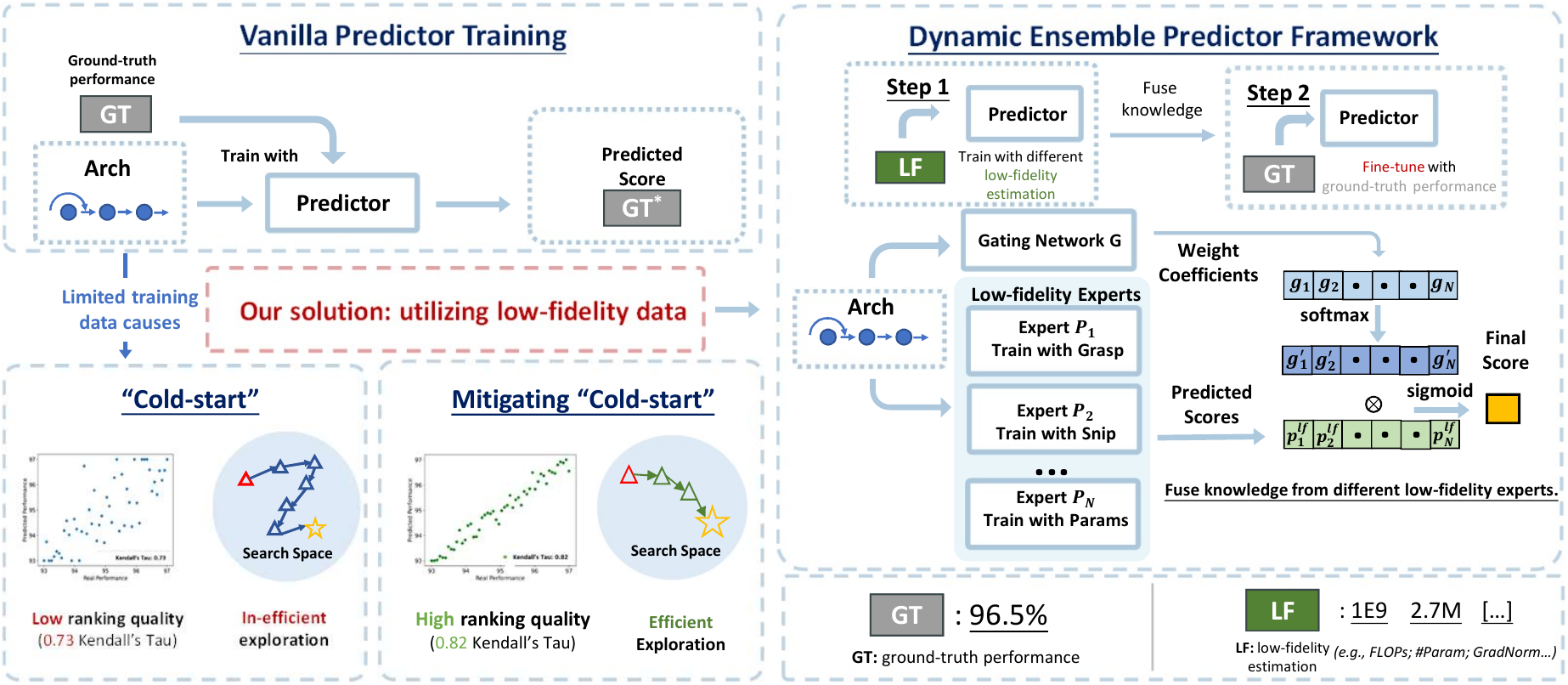}
\caption{Illustration of our motivation and our proposed dynamic ensemble predictor framework.}
\label{fig:flow}
\end{figure*}

\section{The Proposed Method}
In this section, we present the dynamic ensemble performance predictor framework. The illustration of our motivation and the predictor framework are shown in Figure~\ref{fig:flow}.

\subsection{Dynamic Ensemble Performance Predictor}




\subsubsection{Dynamic Ensemble Neural Predictor.} Suppose we have $N$ predictors $\{P_i\}_{i=1}^N$ (i.e., low-fidelity experts), each of which takes the architecture $\alpha$ as the input and outputs a predicted score. 
We represent the predicted scores for the architecture $\alpha$ as 
\begin{equation}
\begin{aligned}
 p^{\mathrm{lf}}(\alpha) = [p^{\mathrm{lf}}_1(\alpha), p^{\mathrm{lf}}_2(\alpha), \cdots, p^{\mathrm{lf}}_N(\alpha)] \in \mathbb{R}^{N}
\end{aligned}
\label{eq:predictor-each}
\end{equation}
where $p^{\mathrm{lf}}_i(\alpha)$ denotes the score predicted by the expert $P_i$.

We learn a gating network $G$ to ensemble these experts to fuse beneficial knowledge from different types of low-fidelity information. The gating network dynamically maps each neural architecture to a set of weights, which are used as the weighting coefficients of predictions from different low-fidelity experts. This enables us to leverage multiple sources of low-fidelity information without worrying about which one is more relevant for the current search space. We utilize the same predictor architecture as the gating network architecture. 

The final predicted score $p(\alpha)$ can be written as
\begin{equation}
\begin{aligned}
&k_{i}(\alpha) = p^{\mathrm{lf}}_i(\alpha) \cdot \frac{\mathrm{exp}{(g_i(\alpha))}}{\sum_{j = 1}^{N}{\mathrm{exp}(g_j(\alpha)})} \\
p(\alpha) =& \mathrm{sigmoid}(G(\alpha)^T p^{\mathrm{lf}}(\alpha)) = \mathrm{sigmoid}(\sum_{i=1}^N k_i(\alpha))
\end{aligned}
\label{eq:predictor-ensemble}
\end{equation}
where $g_i(\alpha)$ and $k_i(\alpha)$ denote the weighting coefficient and weighted score for the $i^{\mathrm{th}}$ low-fidelity expert, and $G(\alpha) \in \mathbb{R}^{N}$ denotes the weighted coefficient vector after softmax. 
The gating network can learn to tailor different weighting coefficients for different input architectures.

\subsubsection{Training Framework.} Our training process for the dynamic ensemble neural predictor consists of two steps. In the first step, we train a predictor on each type of low-fidelity information to extract expert knowledge, as formalized below: 

\begin{equation}
\begin{aligned}
w_i^* = \mathop{\mathrm{argmin}}_{w_i} E_{x^{\mathrm{lf}}\sim D^{\mathrm{lf}}_i}[\mathcal{L}(x^{\mathrm{lf}}, P_i(w_i))]
\end{aligned}
\label{eq:predictor-lf-optim}
\end{equation}
where $x^{\mathrm{lf}}\sim D^{\mathrm{lf}}_i$ denotes the data sampled from the training dataset of the $i^{\mathrm{th}}$ type of low-fidelity information and $w_i$ denotes the weights of the $i^{\mathrm{th}}$ predictor. In the second step, we construct and finetune the entire predictor on the actual performance data, 

\begin{equation}
\begin{aligned}
\{w_i^*\}_{i=1}^N, w_g^* = \mathop{\mathrm{argmin}}_{\{w_i\}_{i=1}^N, w_g} E_{x^{\mathrm{tr}} \sim D^{\mathrm{tr}}}[\mathcal{L}(x^{\mathrm{tr}}, P(\{w_i\}_{i=1}^N, w_g))]
\end{aligned}
\label{eq:predictor-final-optim}
\end{equation}
where $w_g$ denotes the weights of the gating network; $x^{\mathrm{tr}}\sim D^{\mathrm{tr}}$ denotes the data sampled from the training dataset of actual performances.

\subsubsection{Discussion about Simplicity.}
Our method is easily understandable and applicable. Firstly, our method requires no hyper-parameter tuning nor careful low-fidelity information selection. Secondly, our method is general to different search spaces, datasets and encoders, since its two core designs are general instead of specially designed for specific search space properties: 1) utilizing low-fidelity information to improve the prediction ability; 2) dynamically assembling low-fidelity experts to fuse beneficial knowledge from different low-fidelity information. Our method can be applied as long as several low-fidelity information (not necessarily having a good correlation with the actual performance) and an arbitrary architecture encoder are available for the search space.


\begin{table*}[tb]
  \centering
 \resizebox{\linewidth}{!}{
      \begin{tabular}{c|c|c|ccccc}
    \toprule
    \multirow{2}{*}{\textbf{Search Space}} &
	\multirow{2}{*}{\textbf{Encoder}} 
 &\multirow{2}{*}{\textbf{Manner}}&\multicolumn{5}{c}{\textbf{Proportions of training samples}}\\
 \cmidrule(lr){4-8} &
    & & \textbf{1\%}&\textbf{5\%}&\textbf{10\%}&\textbf{50\%} &\textbf{100\%}\\
       \midrule
       \midrule
       \multirow{4}{*}{\textbf{NAS-Bench-201}} &
   \multirow{2}{*}{\textbf{GATES}}
    & Vanilla  & 0.7332$_{(0.0110)}$ & 0.8582$_{(0.0059)}$  & 0.8865$_{(0.0045)}$       & 0.9180$_{(0.0029)}$   & 0.9249$_{(0.0019)}$\\
   & & Ours  & \textbf{0.8244}$_{(0.0081)}$ & \textbf{0.8948}$_{(0.0021)}$ & \textbf{0.9075}$_{(0.0015)}$ & \textbf{0.9216}$_{(0.0019)}$ & \textbf{0.9250}$_{(0.0020)}$\\
   \cmidrule(lr){2-8}
   & \multirow{2}{*}{\textbf{LSTM}}
       & Vanilla  & 0.5692$_{(0.0087)}$ & 0.6410$_{(0.0018)}$  & 0.7258$_{(0.0053)}$ & 0.8765$_{(0.0010)}$   & 0.9000$_{(0.0008)}$\\
   & & Ours  & \textbf{0.7835}$_{(0.0062)}$ & \textbf{0.8538}$_{(0.0029)}$ & \textbf{0.8683}$_{(0.0015)}$ & \textbf{0.8992}$_{(0.0010)}$ & \textbf{0.9084}$_{(0.0010)}$\\
   \midrule
   \multirow{4}{*}{\textbf{NAS-Bench-301}} &
   \multirow{2}{*}{\textbf{GATES}}
    & Vanilla  & 0.4160$_{(0.0450)}$ & 0.6752$_{(0.0088)}$  & 0.7354$_{(0.0044)}$  & 0.7693$_{(0.0041)}$   & \textbf{0.7883}$_{(0.0011)}$\\
   & & Ours  & \textbf{0.5529}$_{(0.0135)}$ & \textbf{0.6830}$_{(0.0038)}$  & \textbf{0.7433}$_{(0.0018)}$ & \textbf{0.7752}$_{(0.0026)}$   & 0.7842$_{(0.0022)}$\\
   \cmidrule(lr){2-8}
   & \multirow{2}{*}{\textbf{LSTM}}
    & Vanilla  & 0.4757$_{(0.0150)}$ & 0.6116$_{(0.0099)}$  & 0.6923$_{(0.0044)}$  & 0.7516$_{(0.0017)}$  & 0.7667$_{(0.0007)}$\\
   & & Ours  & \textbf{0.4805}$_{(0.0083)}$ & \textbf{0.6405}$_{(0.0035)}$  & \textbf{0.7075}$_{(0.0022)}$ & \textbf{0.7544}$_{(0.0028)}$ & \textbf{0.7751}$_{(0.0011)}$\\
   \midrule
   \multirow{2}{*}{\textbf{NDS-ResNet}} &
   \multirow{2}{*}{\textbf{LSTM}}
    & Vanilla  & 0.2549$_{(0.0087)}$ & 0.4564$_{(0.0108)}$  & 0.5770$_{(0.0094)}$  & 0.7758$_{(0.0078)}$   & 0.8244$_{(0.0110)}$\\
   & & Ours  & \textbf{0.7064}$_{(0.0109)}$ & \textbf{0.7548}$_{(0.0080)}$  & \textbf{0.7652}$_{(0.0037)}$ & \textbf{0.8271}$_{(0.0054)}$ & \textbf{0.8383}$_{(0.0049)}$\\
   \midrule
   \multirow{2}{*}{\textbf{NDS-ResNeXt-A}} &
   \multirow{2}{*}{\textbf{LSTM}}
    & Vanilla  & 0.3568$_{(0.0327)}$ & 0.6243$_{(0.0220)}$  & 0.6671$_{(0.0307)}$  & 0.8224$_{(0.0091)}$  & 0.8701$_{(0.0051)}$\\
   & & Ours  & \textbf{0.7753}$_{(0.0010)}$ & \textbf{0.8276}$_{(0.0024)}$ & \textbf{0.8398}$_{(0.0044)}$ & \textbf{0.8453}$_{(0.0040)}$ & \textbf{0.8777}$_{(0.0042)}$\\
   \midrule
   \multirow{2}{*}{\textbf{MobileNet-V3}} &
   \multirow{2}{*}{\textbf{LSTM}}
    & Vanilla  & 0.7373$_{(0.0041)}$ & 0.7852$_{(0.0028)}$  & 0.7832$_{(0.0040)}$  & 0.7944$_{(0.0028)}$   & 0.8023$_{(0.0014)}$\\
   & & Ours  & \textbf{0.7698}$_{(0.0018)}$ & \textbf{0.8034}$_{(0.0027)}$  & \textbf{0.8042}$_{(0.0019)}$ & \textbf{0.8084}$_{(0.0017)}$   & \textbf{0.8135}$_{(0.0020)}$\\
    \bottomrule
    \end{tabular}}
 \caption{The Kendall's Tau (average over five runs) of using different encoders on NAS-Bench-201, NAS-Bench-301, NDS-ResNet, NDS-ResNeXt-A and MobileNet-V3. And the standard deviation is in the subscript. The detailed dataset split is elaborated in the appendix. ``Vanilla'' represents directly training predictor with ground-truth accuracies without low-fidelity information utilization.}
  \label{tab:predictor_kd}
\end{table*}

\subsection{Overall Search Flow}
Our predictor-based flow goes as follows. In the first phase, we mitigate the cold-start issue by utilizing low-fidelity estimation. Specifically, we randomly sample $N_0$  architectures from the search spaces for ground-truth performance and $M$ for low-fidelity information evaluation, respectively. Next, these data are used to train an initial predictor with the dynamic ensemble method. 

In the second phase, we run a predictor-based search for $T_p$ stages. In each stage, a $T_{pe}$-step tournament-based evolutionary search~\cite{real2018regularized} (population size $\pi$, tournament $\mu$) with scores evaluated by the predictor are run for $N_p$ times. We query for rewards of the $N_p$ sampled architectures and then finetune the predictor on all known architecture-performance data for $K$ epochs. In total, we query the actual performance of architectures in search space for $N_0 + T_p \times N_p$ times. The test accuracy of the architecture with the highest reward among all sampled architectures is reported.

Note that our method is a predictor pretraining method that can be easily incorporated into most predictor-based NAS frameworks to alleviate the cold-start problem. It is compatible with different types of predictor architectures~\cite{luo2018neural,ning2020generic,yan2021cate} or search frameworks~\cite{luo2018neural,ning2020generic,shi2020bridging}. 


\section{Experiments and Results}
In this section, we conduct experiments across different search spaces and under various experimental settings to evaluate the dynamic ensemble performance predictor. 


\subsection{Evaluation of Prediction Ability}
\label{sec:predict-ability-evaluation}
To begin with, we evaluate the prediction ability improvement brought by the proposed dynamic ensemble performance predictor on several public benchmarks with different training ratios and architecture encoders. These experiments mainly compare the ranking qualities of predictors. 

\subsubsection{Search Space.} We conduct experiments on the five search spaces for a thorough evaluation: NAS-Bench-201, NAS-Bench-301, NDS-ResNet / ResNeXt-A and MobileNet-V3.


We divide architectures into a training and validation split for each space. We train the predictors on the former and test their prediction ability on the latter. All architectures in the training split with different types of low-fidelity information are used in the first training step. The detailed search space description, data split, types and acquisition of the utilized low-fidelity information are elaborated in the appendix.

\begin{figure*}[t]
\center
\subcaptionbox{\label{fig:search_nb201}}
{\includegraphics[width=0.40\linewidth]{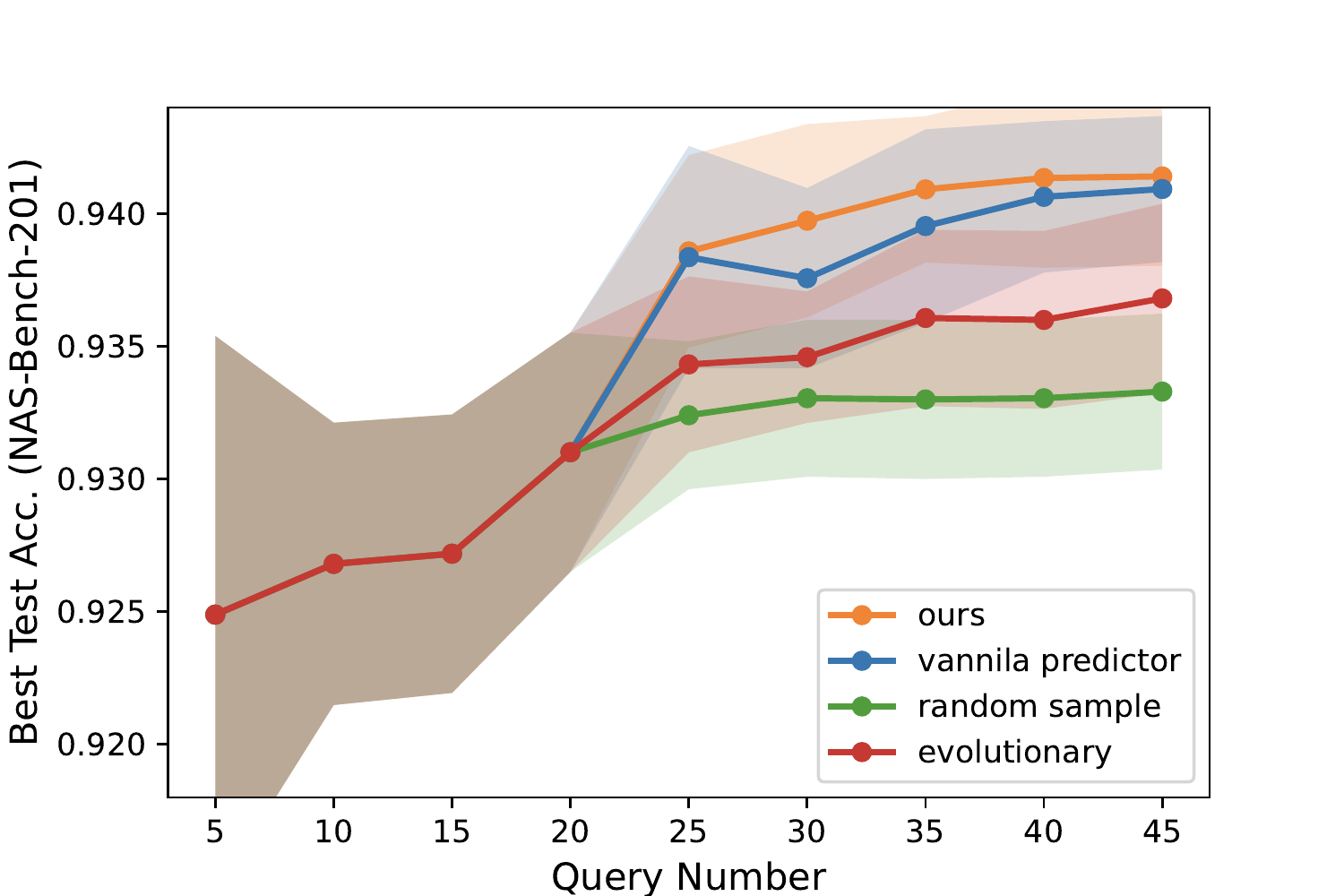}}
\subcaptionbox{\label{fig:search_nb301}}
{\includegraphics[width=0.40\linewidth]{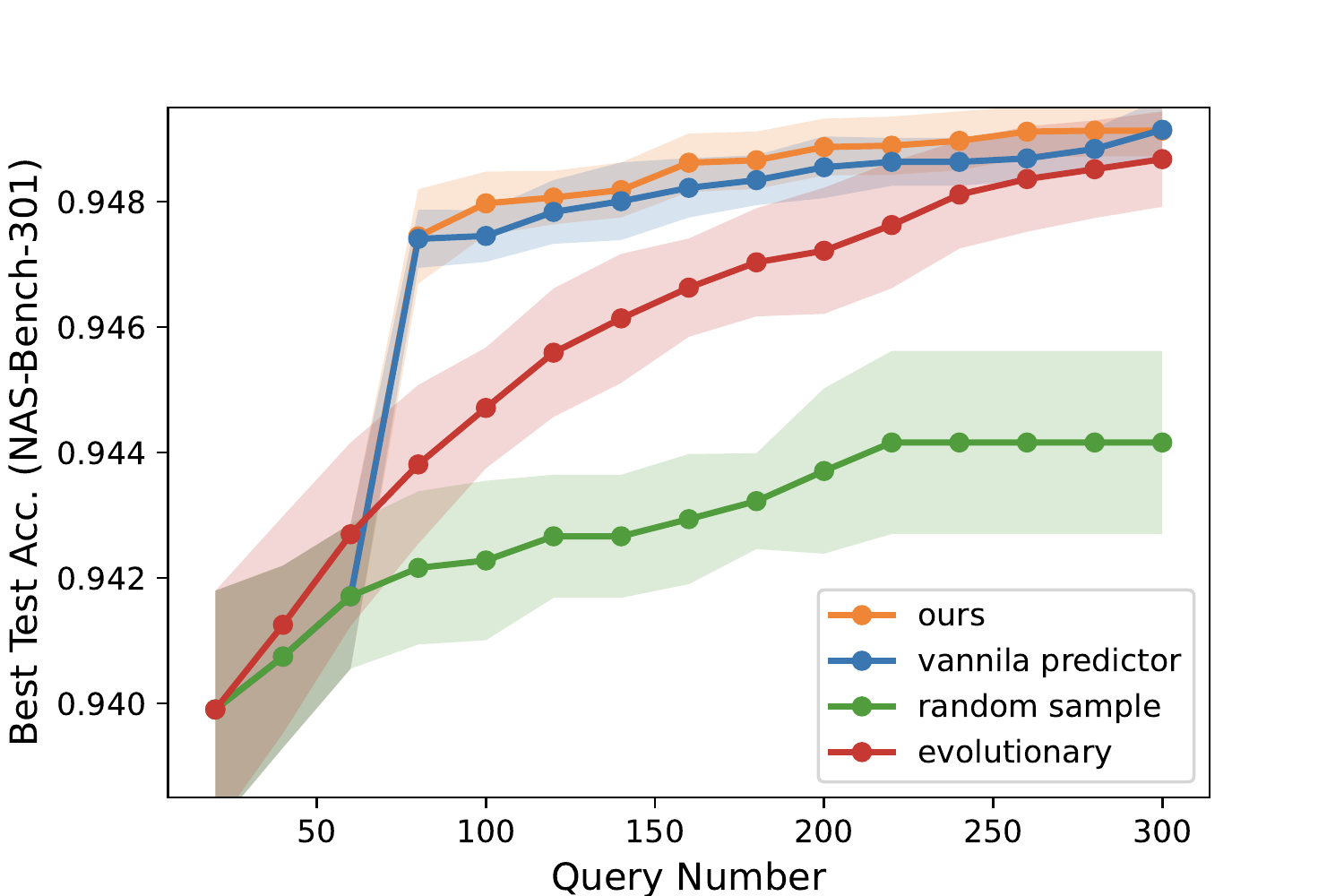}}
\caption{Comparison with other search strategies on NAS-Bench-201 (Figure~\ref{fig:search_nb201}) and NAS-Bench-301 (Figure~\ref{fig:search_nb301}). We report the test accuracy of the architecture with the highest reward among all sampled architectures.}
\label{fig:search_nb}
\end{figure*}

\begin{table*}[!t]
  \centering
 \resizebox{\linewidth}{!}{
  \begin{tabular}{c|c|cc|cc|cc}
    \toprule
	\multirow{2}{*}{\textbf{Method}} &\multirow{2}{*}{\textbf{search seconds}}
	& \multicolumn{2}{c|}{\textbf{CIFAR-10 (\%)}} & \multicolumn{2}{c|}{\textbf{CIFAR-100 (\%)}} & \multicolumn{2}{c}{\textbf{ImageNet-16-120 (\%)}}\\
    \cline{3-8}
    & & valid & test & valid & test & valid & test\\
	\hline 
	\midrule
   RSPS~\cite{li2019random} & 7587.12 &  84.16$_{(1.69)}$ & 87.66$_{(1.69)}$ & 59.00$_{(4.60)}$ & 58.33$_{(4.34)}$ & 31.56$_{(3.28)}$ & 31.14$_{(3.88)}$ \\
   DARTS-V2~\cite{liu2018darts} & 29901.67 & 39.77$_{(0.00)}$ & 54.30$_{(0.00)}$ & 15.03$_{(0.00)}$ & 15.61$_{(0.00)}$ & 16.43$_{(0.00)}$ & 16.32$_{(0.00)}$ \\
   GDAS~\cite{dong2019GDAS} &  28925.91 & 90.00$_{(0.21)}$ & 93.51$_{(0.13)}$ & 71.15$_{(0.27)}$ & 70.61$_{(0.26)}$ & 41.70$_{(1.26)}$ & 41.84$_{(0.90)}$ \\
   SETN~\cite{dong2019os} & 31009.81 &  82.25$_{(5.17)}$ & 86.19$_{(4.63)}$ & 56.86$_{(7.59)}$ & 56.87$_{(7.77)}$ & 32.54$_{(3.63)}$ & 31.90$_{(4.07)}$ \\
   ENAS-V2~\cite{pham2018efficient} & 13314.51 & 39.77$_{(0.00)}$ & 54.30$_{(0.00)}$ & 15.03$_{(0.00)}$ & 15.61$_{(0.00)}$ & 16.43$_{(0.00)}$ & 16.32$_{(0.00)}$ \\
   \hline
   \midrule
   Random Sample & 0.01 & 90.03$_{(0.36)}$ & 93.70$_{(0.36)}$ & 70.93$_{(1.09)}$ & 71.04$_{(1.07)}$ & 44.45$_{(1.10)}$ & 44.57$_{(1.25)}$ \\
   NPENAS~\cite{wei2020npenas} & - & 91.08$_{(0.11)}$ & 91.52$_{(0.16)}$ & - & - & - & - \\
   REA~\cite{real2018regularized} & 0.02 & 91.19$_{(0.31)}$ & 93.92$_{(0.30)}$ & 71.81$_{(1.12)}$ & 71.84$_{(0.99)}$ & 45.15$_{(0.89)}$ & 45.54$_{(1.03)}$ \\
   NASBOT~\cite{white20219study} 
   & -    &  - & 93.64$_{(0.23)}$ & - & 71.38$_{(0.82)}$ & - & 45.88$_{(0.37)}$ \\
   REINFORCE~\cite{williams1992Simple} & 0.12 &  91.09$_{(0.37)}$ & 93.85$_{(0.37)}$ & 71.61$_{(1.12)}$ & 71.71$_{(1.09)}$ & 45.05$_{(1.02)}$ & 45.24$_{(1.18)}$ \\
   BOHB~\cite{falkner2018bohb} & 3.59 &  90.82$_{(0.53)}$ & 93.61$_{(0.52)}$ & 70.74$_{(1.29)}$ & 70.85$_{(1.28)}$ & 44.26$_{(1.36)}$ & 44.42$_{(1.49)}$ \\
   ReNAS~\cite{xu2021renas} 
   & 86.31 & 90.90$_{(0.31)}$ & 93.99$_{(0.25)}$ & 71.96$_{(0.99)}$ & 72.12$_{(0.79)}$ & 45.85$_{(0.47)}$ & 45.97$_{(0.49)}$ \\
   	\midrule
   Ours & $\sim $720 & \textbf{91.59}$_{(0.02)}$ 
   & \textbf{94.37}$_{(0.00)}$ &\textbf{73.49}$_{(0.00)}$ & \textbf{73.50}$_{(0.00)}$  &\textbf{46.41}$_{(0.06)}$  & \textbf{46.39}$_{(0.01)}$  \\
   \hline
   \midrule
   Optimal & - & 91.61 & 94.37 & 73.49 & 73.51 & 46.77 & 47.31 \\
   \cline{1-1}\cline{2-8}
   \midrule
   ResNet  & - & 90.83 & 93.97 & 70.42 & 70.86 & 44.53 & 43.63 \\ 
   \bottomrule
\end{tabular}
}
  \caption{Search results on NAS-Bench-201. The standard deviation is in the subscript.}
 \label{tab:nb201_search_result}
\end{table*}

\subsubsection{Predictor Construction.}
In the basic predictor prediction flow, an encoder first encodes the architecture into an embedding vector. Then the vector is fed into an MLP to get the prediction score. We use LSTM~\cite{luo2018neural} and GATES~\cite{ning2020generic} as the encoder. 
We only use LSTM for the non-topological search spaces, including NDS-ResNet / ResNeXt-A and MobileNet-V3, since GATES is specially designed for topological architectures. 

\subsubsection{Training Settings.} Following the previous studies~\cite{ning2020generic,xu2021renas}, we train predictors with the hinge pair-wise ranking loss with margin $m=0.1$. We first train different low-fidelity experts for 200 epochs and then finetune the dynamic ensemble performance predictor on the actual performance data for 200 epochs. For comparison, we directly train the vanilla predictor on the actual performance data for 200 epochs. An Adam optimizer with learning rate 1e-3 is applied for optimization. The batch sizes used for NAS-Bench-201, NAS-Bench-301, NDS and MobileNetV3 search spaces are 512, 128, 128 and 512, respectively. 

\subsubsection{Results.} Following previous studies~\cite{ning2020surgery}, we adopt Kendall's Tau (KD)
as the evaluation criteria. As the results shown in Table~\ref{tab:predictor_kd}, our proposed method outperforms \emph{vanilla} predictor training consistently on different search spaces, architecture encoders, and training ratios. Especially, our method brings a larger improvement when the training ratio is smaller. For example, on NDS-ResNet and NDS-ResNeXt-A, our method achieves 0.7064 and 0.7753 Kendall's Tau with 1\% training samples, respectively, much better than the vanilla predictor (0.2549, 0.3568). 


\subsection{Mitigating the Cold-Start Issue}
\label{tab:search-evaluation}
We conduct architecture search on several search spaces to demonstrate that our method can effectively mitigate the cold-start issue and boost the performance of NAS. 

\subsubsection{Search on NAS-Bench-201.}
We conduct experiments on NAS-Bench-201 under three settings with architectures encoded by GATES. We use the validation accuracy on the CIFAR-10 dataset as the reward to guide the search. 

\paragraph{\textbf{Comparison with Different Search Strategies.}}
\label{sec:nb201_search_setting}
We compare our method with random sample, tournament-based evolutionary ($\pi=20,\mu=5$), and predictor-based flow without utilizing low-fidelity information. Each method is run ten times. We set $N_0=20$, $M=7813$, $T_p=5$, $T_{pe}=50$, $N_p=5$, $\pi=20$, $\mu=5$ and $K=100$ for our method. In total, we query the benchmark 45 times and report the test accuracy of the best architecture selected by the predictor. As the results shown in Figure~\ref{fig:search_nb201}, the best architectures discovered by our method have higher test accuracies using the same query times.

\begin{table*}[!ht]
  \centering
 \resizebox{\linewidth}{!}{
  \begin{tabular}{c|ccc|c|c|c}
    \toprule
	\multirow{2}{*}{\textbf{Method}} 
	& \multicolumn{3}{c|}{\textbf{Test Error (\%)}} & \multicolumn{1}{c|}{\textbf{Param}$\dagger$} & \multicolumn{1}{c|}{\textbf{FLOPs}$\dagger$} & \multicolumn{1}{c}{\textbf{Architecture}} \\
    & \textbf{CIFAR-10} & \textbf{CIFAR-100} & \textbf{ImageNet (Top1/Top5)}  & \textbf{(M)} &\textbf{(M)} & \textbf{Optimization}\\
	\midrule
	\midrule
   NASNet-A~\cite{zoph2018learning} &2.65 &17.81 &26.0 / 8.4 & 3.3 &564 &RL\\
   PNAS~\cite{liu2019PNAS} &3.41$\pm$0.09 &17.63 &25.8 / 8.1 & 3.2 &588 &SMBO\\
   EN$^2$AS~\cite{zhang2020differentiable} &2.61$\pm$0.06 &16.45 &26.7 / 8.9 &3.1 &506 &EA\\
   NSAS~\cite{zhang2020overcoming} &2.59$\pm$0.06 &17.56 &25.5 / 8.2 &3.1 &506 & random\\
   \midrule
   \midrule
   DARTS~\cite{liu2018darts} &2.76$\pm$0.09 &17.54 &26.9 / 8.7 &3.4 &574 & gradient\\
   GDAS~\cite{dong2019GDAS} &2.93&18.38&26.0 / 8.5 & 3.4 & 545 &gradient\\
   SNAS~\cite{xie2018SNAS} &2.85$\pm$0.02 &20.09 &27.3 / 9,2 &2.8 &474 & gradient\\
   PC-DARTS~\cite{xu2019pc} &2.57$\pm$0.03 &17.11 &25.1 / 7.8 &3.6 &586 &gradient\\
   \midrule
   \midrule
   NAO~\cite{luo2018neural} & 2.48 & 15.67$\ddagger$ & 25.7 / 8.2 &   10.6 & 584 & Predictor-based\\
   GATES~\cite{ning2020generic} & 2.58 & - & - & 4.1 & - & Predictor-based\\
   BANANAS~\cite{white2019bananas} &2.57 &- & - & 4.0 & - & Predictor-based\\
   NPENAS-BO~\cite{wei2020npenas} &2.64 $\pm$ 0.08 & - & - &- & - & Predictor-based\\
   NAS-BOWL~\cite{ru2021interpretable} &2.61 $\pm$ 0.08 & - & - &3.7 & - & Predictor-based\\
   \midrule
   \midrule
   Ours &  \textbf{2.30} &\textbf{16.07} &\textbf{24.4} / \textbf{7.4} &4.1 &645 &Predictor-based \\
   \bottomrule
\end{tabular}
}
\begin{minipage}{0.95\textwidth}
    {\small
    $\dagger$: ``Param'' is the model size of CIFAR-10 model, while ``FLOPs'' is calculated based on the ImageNet models.
    
      $\ddagger$: This architecture is much larger than ours.
      }
    \end{minipage}
  \caption{Test error comparison with other NAS methods on CIFAR-10, CIFAR-100, and ImageNet.}
 \label{tab:darts_final_result}
\end{table*}

\begin{table*}[!t]
\centering
\resizebox{0.56\linewidth}{!}{
\begin{tabular}{c|ccc|c}
\toprule
\textbf{Search Space}
&\textbf{Vanilla} &\textbf{Ours} &\textbf{Random Sample} &\textbf{Optimal}\\
\midrule
\midrule
\textbf{NDS-ResNet} & 0.9437$_{(0.0000)}$ & \textbf{0.9488$_{(0.0001)}$} & 0.9420$_{(0.0021)}$ & 0.9516 \\
\textbf{NDS-ResNeXt-A} & 0.9454$_{(0.0005)}$ & \textbf{0.9456$_{(0.0000)}$} & 0.9368$_{(0.0038)}$ & 0.9483 \\
\textbf{MobileNet-V3} & 0.7718$_{(0.0000)}$ & \textbf{0.7721$_{(0.0002)}$} & 0.7633$_{(0.0034)}$ & 0.7749 \\
\bottomrule
\end{tabular}
}
\caption{Discovered Architecture accuracies. We report the average values and the standard deviation is in the subscript. 
}
\label{tab:other-search-constrains}
\end{table*}

\paragraph{\textbf{Comparison with AceNAS and ProxyBO.}}
We compare our method with ProxyBO~\cite{shen2021proxybo} and AceNAS~\cite{zhang2021acenas} to verify that our method is a better way to 
utilize cheaper-to-obtain estimations. 
All search settings are kept the same as in other experiments, except for $N_p=20$ and $T_p = 9$. In total, we query the benchmark 200 times and report the accuracy of the best architecture selected by the predictor. The search budget is the same as that of ProxyBO but less than AceNAS (500 queries). We run our method ten times with different seeds. Our method gets an 8.39\% validation error on CIFAR-10 and 26.50\% test error on CIFAR-100, respectively, better than ProxyBO (8.56\%, 26.53\%) and AceNAS (26.62\% on CIFAR-100).

\paragraph{\textbf{Comparison with Other NAS Methods.}}
For a fair comparison with other NAS methods, following ~\cite{xu2021renas}, we finetune the predictor in the second step with 90 randomly sampled architectures and their corresponding rewards. Then we traverse the search space with the predictor and report the best validation and test accuracies among the top-10 architectures selected by the predictor. We run the experiment 10 times and report the average and standard values\footnote{Following ReNAS~\cite{xu2021renas}, we report the predictor training time as the search cost.}. As shown in Table~\ref{tab:nb201_search_result}, our method achieves better performance than the other methods on all three datasets. Remarkably, by utilizing one-shot estimation in predictor-based NAS, our method significantly outperforms the original one-shot method ENAS \cite{pham2018efficient}.


\subsubsection{Search on NAS-Bench-301.}  With GATES as the encoder, we set $N_0=60$, $M=5896$, $T_p=10$, $T_{pe}=100$, $N_p=20$, $\pi=20$, $\mu=10$ and $K=100$ for our method. In total, we query the benchmark 300 times. We compare our methods with random sample, tournament-based evolutionary ($\pi=20$, $\mu=10$), and the same predictor-based flow but without the utilization of low-fidelity information. Each method is run ten times with different seeds. The result is shown in Figure~\ref{fig:search_nb301}. Our method achieves better test accuracies than predictor-based flow without the utilization of low-fidelity information and outperforms random sample and evolutionary methods by a large margin.

\subsubsection{Search on DARTS.}
We further employ our method in the DARTS~\cite{liu2018darts} search space. For fast experiments, we conduct architecture search on the NAS-Bench-301 benchmark, which is similar to the DARTS space. The search settings are the same as those on NAS-Bench-301. In total, we query the benchmark 300 times. After the search process, the best-discovered architecture is augmented following the DARTS setting and trained from scratch to get the final test accuracy. Detailed discovered architecture training settings are elaborated in the appendix. The comparison of the test errors is shown in Table~\ref{tab:darts_final_result}. Our method achieves a test error of 2.30\% on CIFAR-10, better than the previous one-shot NAS methods, such as DARTS (3.00\%) and GDAS (2.93\%). When transferred to CIFAR-100 and ImageNet, the discovered architecture achieves a test error of 16.07\% and 24.4\%, respectively, also outperforming the other architectures.

\subsubsection{Search on MobileNet-V3 and NDS.}
After the first training step, we finetune the entire predictor on 1\% architectures in the training split. On MobileNet-V3, we traverse 100000 randomly sampled architectures with the predictor. On NDS-ResNet / ResNeXt-A, we traverse all the architectures in the search space. The best test accuracy among the top-10 architectures selected by the predictor is reported. As shown in Table~\ref{tab:other-search-constrains}, compared with other strategies, our method consistently discovers superior architectures.

\subsubsection{Efficiency Comparison with Other Methods.}
Except for the one-shot score, most types of low-fidelity information can be obtained at an extremely low cost. For example, evaluation of the parameter size for all architectures in NAS-Bench-201 can be accomplished within a minute. Although utilizing the one-shot information involves supernet training and submodel testing, our method is still more efficient than baselines. For example, the cost of training the supernet and testing 7813 candidate architectures on NAS-Bench-201 is comparable to training about 15 architectures for 200 epochs. In our experiment, when querying 45 architectures to get their ground-truth performance in NAS-Bench-201, the equivalent total cost is about training 45+15=60 architectures. And the accuracy of our discovered architecture (94.09\%) surpasses the architecture accuracy (93.99\%) discovered by ReNAS after 90 queries by a large margin.

 


\begin{figure}[!t]
\center
\includegraphics[width=\linewidth]{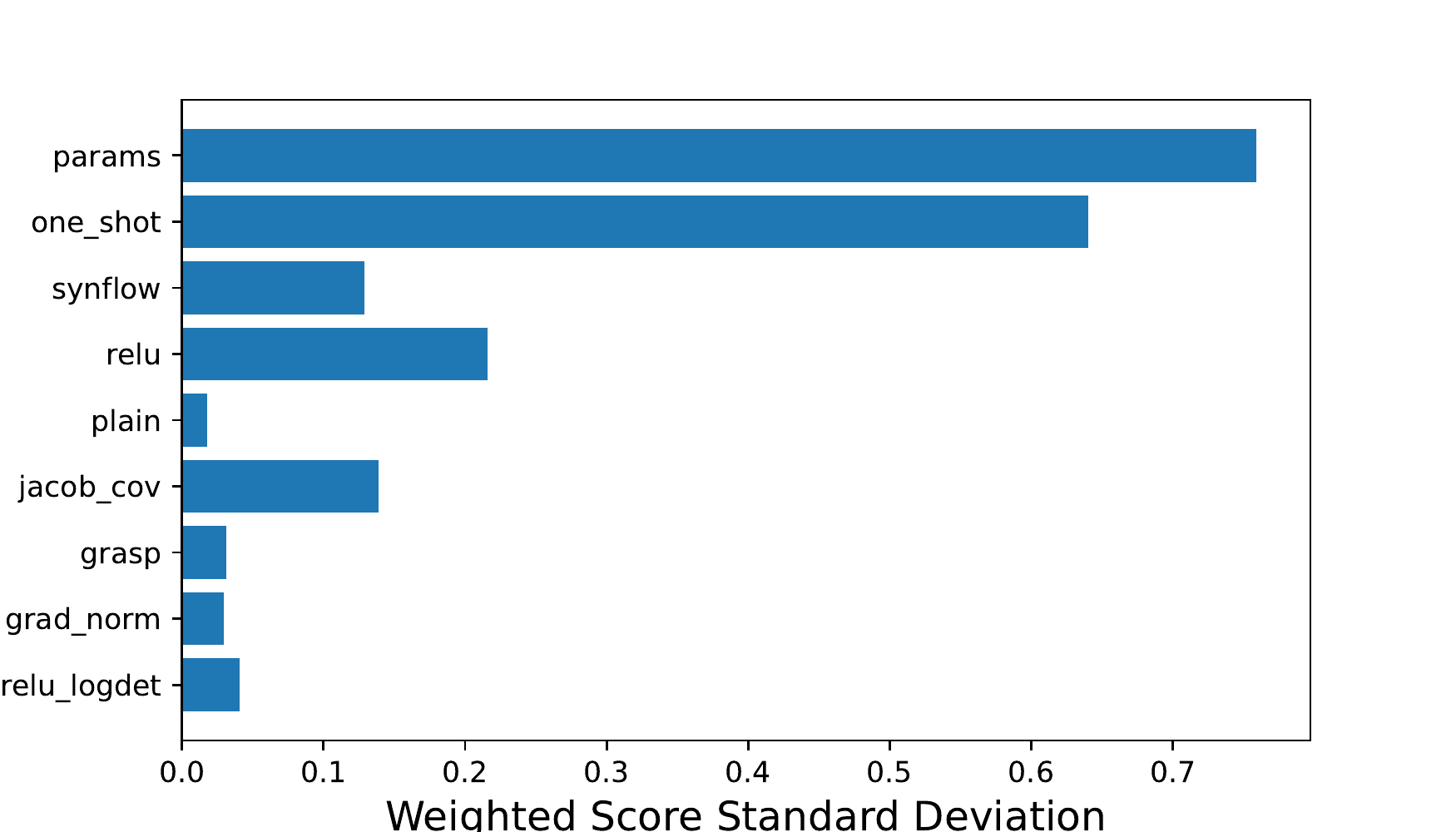}
\caption{Standard deviation of weighted scores of different low-fidelity experts on NDS-ResNet.}
\label{fig:error_bar}
\end{figure}


\subsection{Empirical Analysis}
\subsubsection{Dynamic Ensemble Analysis.} Since we model predictor learning as a ranking problem, the absolute value of the weighted score by an expert does not reflect its importance directly. This is because the output range of experts varies. If there is a low-fidelity expert that has the highest weighting coefficient, whose weighted scores for all architectures are the same. Then, this expert does not contribute new information to the architecture ranking. In other words, only the difference between weighted scores of architectures by an expert contributes to the architecture ranking. So, instead, we calculate the standard deviation of the weighted scores $k_i$ as the criterion. Figure~\ref{fig:error_bar} shows the results of different low-fidelity experts on NDS-ResNet. The experts on the parameter size and one-shot score have a much larger standard deviation than other types, indicating that the predictor highly relies on them for prediction.
On the other hand, in the appendix, we empirically verify that these two low-fidelity information types are the most beneficial ones on NDS-ResNet. That is to say, the relative importance of low-fidelity experts in our predictor aligns well with the extent of benefits brought by low-fidelity information when only one type of information is used. This backs the rationality of using our method to automatically and adequately combine different low-fidelity information.




\subsubsection{Comparison with Uniform Weight Learning.} An alternative to the dynamic ensemble is to learn a uniform set of coefficients for all the architecture in the search space. However, considering different types of low-fidelity information have different prediction abilities for different regions of the search space, the weighting coefficients for architectures would better be different. To verify this intuition, we conduct experiments on NAS-Bench-201 and NDS ResNet / ResNeXt-A with the LSTM encoder for comparison. In addition, we also make comparison with a simple ensemble method that just averages outputs of different low-fidelity experts. As shown in Table~\ref{tab:compare-union}, the dynamic ensemble method consistently outperforms the uniform and simple ensemble method, demonstrating the effectiveness of our method.

\begin{table}[tb]
\centering
\resizebox{\linewidth}{!}{
\begin{tabular}{c|c|ccc}
\toprule
\multirow{2}{*}{\textbf{Search Space}} &\multirow{2}{*}{\textbf{Manner}}&\multicolumn{3}{c}{\textbf{Proportions of training samples}}\\
\cmidrule(lr){3-5} & & \textbf{1\%}&\textbf{5\%}&\textbf{10\%}\\
\midrule
\midrule
\multirow{3}{*}{\textbf{NAS-Bench-201}}
& Simple & 0.6936$_{0.0038}$ & 0.7763$_{(0.0058)}$ & 0.8218$_{(0.0015)}$ \\
& Uniform  & 0.7442$_{(0.0031)}$ & 0.8296$_{(0.0019)}$  & 0.8549$_{(0.0004)}$  \\
& Ours  & \textbf{0.7835}$_{(0.0062)}$ & \textbf{0.8538}$_{(0.0029)}$ & \textbf{0.8683}$_{(0.0015)}$\\
\midrule
\multirow{3}{*}{\textbf{NDS ResNet}} 
& Simple & 0.5789$_{(0.0145)}$ & 0.7247$_{(0.0088)}$ & 0.7349$_{(0.0125)}$ \\
& Uniform  & 0.6794$_{(0.0174)}$ & 0.7302$_{(0.0055)}$  & 0.7452$_{(0.0052)}$  \\
& Ours  & \textbf{0.7064}$_{(0.0109)}$ & \textbf{0.7548}$_{(0.0080)}$ & \textbf{0.7652}$_{(0.0037)}$\\
\midrule
\multirow{3}{*}{\textbf{NDS ResNeXt-A}}
& Simple & 0.7326$_{(0.0122)}$ & 0.7942$_{(0.0073)}$ & 0.8009$_{(0.0042)}$ \\
& Uniform  & 0.7694$_{(0.0062)}$ & 0.8253$_{(0.0033)}$  & 0.8348$_{(0.0040)}$  \\
& Ours  & \textbf{0.7753}$_{(0.0010)}$ & \textbf{0.8276}$_{(0.0024)}$ & \textbf{0.8398}$_{(0.0044)}$\\
\bottomrule
\end{tabular}}
 \caption{The Kendall's Tau (average over five runs) of using the LSTM encoder on NAS-Bench-201 and NDS-ResNet / ResNeXt-A. The standard deviation is in the subscript. ``Uniform'' represents learning a uniform set of weight coefficients for all the architectures. ``Simple'' represents simply averaging outputs of different low-fidelity experts.}
  \label{tab:compare-union}
\end{table}

\section{Conclusion}
\label{sec:conc}
This paper proposes to leverage low-fidelity information to mitigate the ``cold-start'' problem of predictor-based NAS. 
Despite the intuitiveness of this idea, we observe that utilizing inappropriate low-fidelity information might damage the prediction ability and different search spaces have different preferences for the utilized low-fidelity information type. 
To circumvent the need to manually decide on which low-fidelity information to use for each architecture and search space, we propose a dynamic ensemble predictor framework to fuse beneficial information from different low-fidelity experts automatically.
Experiments across five search spaces with different architecture encoders under various experimental settings demonstrate the effectiveness of our methods. Our method can be easily incorporated with existing predictor-based NAS methods to boost search performances.

\section{Acknowledgments}
This work was supported by National Natural Science Foundation of China (No. U19B2019, 61832007), Huawei Noah's Ark, Beijing National Research Center for Information Science and Technology (BNRist), Tsinghua EE Xilinx AI Research Fund, and Beijing Innovation Center for Future Chips. We thank the anonymous reviewers for their constructive suggestions.

\clearpage
\bibliography{aaai23}

\clearpage
\begin{appendices}
\renewcommand{\thefigure}{A\arabic{figure}}
\setcounter{figure}{0}
\renewcommand{\thetable}{A\arabic{table}}
\setcounter{table}{0}
\setcounter{secnumdepth}{2} 

\section{Additional Experiments}
\subsection{Detailed Preliminary Experiment}
\begin{figure}[!t]
\center
\includegraphics[width=\linewidth]{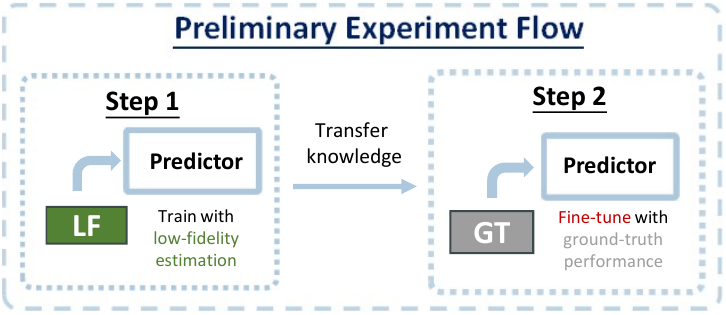}
\caption{Illustration of the preliminary low-fidelity information utilization flow. We pretrain the predictor on a single type of low-fidelity information and then finetune the predictor on the actual architecture-performance data.}
\label{fig:preliminary-flow}
\end{figure}

Inspired by recent advances in transfer learning, we can simply model the utilization of low-fidelity information as a knowledge transfer problem from the low-fidelity task (source domain) to actual performance prediction (target domain). Specifically, we pretrain the predictor on a single type of low-fidelity information data and then finetune it on a small amount of actual architecture-performance data. The illustration of the preliminary flow is shown in Figure~\ref{fig:preliminary-flow}.

We conduct this preliminary experiment across five search spaces, utilizing different types of low-fidelity information. We report the average Kendall's Tau and make a comparison with the vanilla predictor training method as well as our proposed dynamic ensemble method in Table~\ref{tab:detailed-nds-mobilenet} (NDS-ResNet / ResNeXt-A~\cite{radosavovic2019nds} and MobileNet-V3~\cite{cai2019once}) and Table~\ref{tab:detailed-nasbench} (NAS-Bench-201~\cite{dong2020bench} and NAS-Bench-301~\cite{siems2020bench}). 

\subsubsection{Observations.} On the one hand, low-fidelity information does have the potential to improve prediction ability with limited actual architecture-performance data significantly. For example, with 1\% training samples, utilizing grad\_norm~\cite{abdelfattah2021zero} increases the Kendall's Tau from 0.2549 to 0.4661 on NDS-ResNet; utilizing synflow~\cite{tanaka2020pruning} increases the Kendall's Tau from 0.3568 to 0.7281 on NDS-ResNeXt-A; utilizing the number of floating-point operations (FLOPs) increases the Kendall's Tau from 0.4160 to 0.5392 on NAS-Bench-201.

On the other hand, inappropriate types of low-fidelity information even damage the prediction ability. For example, with 1\% training samples, utilizing relu~\cite{ning2020surgery} decreases the Kendall's Tau from 0.3568 to 0.3062 on MobileNet-V3; utilizing plain~\cite{1988Skeletonization} decreases the Kendall's Tau from 0.5692 to 0.4882 on NAS-Bench-201; utilizing synflow\_bn~\cite{tanaka2020pruning} decreases the Kendall's Tau from 0.4160 to 0.2197 on NAS-Bench-301.

The above observations inspire us to discover a type of low-fidelity information that is consistently effective in different search spaces. Unfortunately, different search spaces have different preferences for low-fidelity information types. For example, utilizing jacob\_cov~\cite{Mellor2020NASwithoutTraining} improves the ranking quality on NAS-Bench-201 but consistently decreases the Kendall's Tau on NDS-ResNet with different proportions of training samples. This phenomenon is understandable. Since the properties of different search spaces are different, the effectiveness of utilizing low-fidelity information will naturally differ. 

Meanwhile, the effectiveness of utilizing a kind of low-fidelity information is not only related to the search space but also the proportion of training samples. For example, utilizing fisher~\cite{Theis2018faster} is beneficial with 1\% and 50\% training samples, but damages or has no impact on performance with the other training sample proportions. Therefore, it is not easy to find or design a type of low-fidelity information that is always effective in different search spaces and different amounts of training data.

\begin{table*}[!t]
\centering
\resizebox{\linewidth}{!}{
\begin{tabular}{c|c|c|ccccc}
\toprule
\multirow{2}{*}{\textbf{Search Space}}
&\multirow{2}{*}{\textbf{Low-fidelity}} & \multirow{2}{*}{\textbf{Corr.}} &\multicolumn{5}{c}{\textbf{Proportions of training samples}}\\
\cmidrule(lr){4-8} &&& \textbf{1\%}&\textbf{5\%}&\textbf{10\%}&\textbf{50\%}&\textbf{100\%}\\
\midrule
\midrule
\multirow{11}{*}{\textbf{NDS-ResNet}}
& \textbf{grad\_norm} & 0.2372 & 0.4661$_{(0.0230)}$ & 0.6634$_{(0.0088)}$  & 0.7267$_{(0.0045)}$ & 0.8138$_{(0.0125)}$ & 0.8292$_{(0.0106)}$\\
& \textbf{grasp} & -0.1142 & 0.0145$_{(0.0145)}$ & 0.3819$_{(0.0336)}$  & 0.4922$_{(0.0239)}$ & 0.7317$_{(0.0135)}$ & 0.8072$_{(0.0085)}$\\
& \textbf{jacob\_cov} & -0.0724 & 0.1134$_{(0.0164)}$ & 0.3374$_{(0.0123)}$  & 0.4234$_{(0.0234)}$ & 0.6824$_{(0.0063)}$ & 0.7669$_{(0.0161)}$\\
& \textbf{one-shot} & 0.6658 & 0.6296$_{(0.0083)}$ & 0.7469$_{(0.0097)}$  & \textbf{0.7779$_{(0.0123)}$} & 0.8238$_{(0.0132)}$ & 0.8354$_{(0.0106)}$\\
& \textbf{parameter-size} & 0.5998 & 0.6737$_{(0.0068)}$ & \textbf{0.7592$_{(0.0052)}$}  & 0.7681$_{(0.0063)}$ & 0.8108$_{(0.0057)}$ & 0.8359$_{(0.0082)}$\\
& \textbf{plain} & 0.3066 & 0.4455$_{(0.0027)}$ & 0.6222$_{(0.0168)}$  & 0.6782$_{(0.0081)}$ & 0.7770$_{(0.0104)}$ & 0.8169$_{(0.0065)}$\\
& \textbf{relu} & 0.0733 & 0.2057$_{(0.0201)}$ & 0.5051$_{(0.0404)}$  & 0.6337$_{(0.0227)}$ & 0.8000$_{(0.0130)}$ & 0.8333$_{(0.0104)}$\\
& \textbf{relu\_logdet} & 0.1817 & 0.3128$_{(0.0123)}$ & 0.4997$_{(0.0115)}$  & 0.5783$_{(0.0049)}$ & 0.7642$_{(0.0174)}$ & 0.8128$_{(0.0189)}$\\
& \textbf{synflow} & 0.2307 & 0.4402$_{(0.0153)}$ & 0.7133$_{(0.0087)}$  & 0.7273$_{(0.0085)}$ & 0.8061$_{(0.0133)}$ & 0.8363$_{(0.0104)}$\\
\cmidrule(lr){2-8}
& \textbf{vanilla} & - & 0.2549$_{(0.0087)}$ & 0.4564$_{(0.0108)}$  & 0.5770$_{(0.0094)}$ & 0.7758$_{(0.0078)}$ & 0.8244$_{(0.0110)}$\\
& \textbf{ours} & - & \textbf{0.7064$_{(0.0109)}$} & 0.7548$_{(0.0080)}$  & 0.7652$_{(0.0037)}$ & \textbf{0.8271$_{(0.0054)}$} & \textbf{0.8383$_{(0.0049)}$}\\
\midrule
\multirow{11}{*}{\textbf{NDS-ResNeXt-A}}
& \textbf{grad\_norm} & 0.3190 & 0.5797$_{(0.0226)}$ & 0.7092$_{(0.0087)}$  & 0.7315$_{(0.0165)}$ & 0.8493$_{(0.0083)}$ & 0.8737$_{(0.0047)}$\\
& \textbf{grasp} & -0.2615 & -0.0082$_{(0.0157)}$ & 0.4016$_{(0.0335)}$  & 0.5289$_{(0.0166)}$ & 0.7983$_{(0.0127)}$ & 0.8543$_{(0.0067)}$\\
& \textbf{jacob\_cov} & 0.0510 & 0.1541$_{(0.0077)}$ & 0.4123$_{(0.0232)}$  & 0.4967$_{(0.0212)}$ & 0.7553$_{(0.0180)}$ & 0.8457$_{(0.0133)}$\\
& \textbf{one-shot} & 0.7221 & \textbf{0.7808$_{(0.0033)}$} & 0.8188$_{(0.0009)}$  & 0.8378$_{(0.0019)}$ & \textbf{0.8517$_{(0.0070)}$} & 0.8716$_{(0.0028)}$\\
& \textbf{parameter-size} & 0.6683 & 0.6871$_{(0.0097)}$ & 0.7429$_{(0.0100)}$  & 0.7668$_{(0.0052)}$ & 0.8273$_{(0.0082)}$ & 0.8715$_{(0.0049)}$\\
& \textbf{plain} & 0.2887 & 0.5018$_{(0.0232)}$ & 0.6714$_{(0.0114)}$  & 0.7025$_{(0.0189)}$ & 0.8431$_{(0.0073)}$ & 0.8710$_{(0.0042)}$\\
& \textbf{relu} & 0.2590 & 0.3062$_{(0.0719)}$ & 0.6438$_{(0.0286)}$  & 0.6819$_{(0.0348)}$ & 0.8495$_{(0.0068)}$ & 0.8740$_{(0.0008)}$\\
& \textbf{relu\_logdet} & 0.4589 & 0.5609$_{(0.0108)}$ & 0.6871$_{(0.0205)}$  & 0.7159$_{(0.0181)}$ & 0.8106$_{(0.0078)}$ & 0.8744$_{(0.0038)}$\\
& \textbf{synflow} & 0.6904 & 0.7281$_{(0.0031)}$ & 0.8016$_{(0.0036)}$  & 0.8187$_{(0.0040)}$ & 0.8370$_{(0.0039)}$ & 0.8743$_{(0.0054)}$\\
\cmidrule(lr){2-8}
& \textbf{vanilla} & - & 0.3568$_{(0.0327)}$ & 0.6243$_{(0.0220)}$  & 0.6671$_{(0.0307)}$ & 0.8224$_{(0.0091)}$ & 0.8701$_{(0.0051)}$\\
& \textbf{ours} & - & 0.7753$_{(0.0010)}$ & \textbf{0.8276$_{(0.0024)}$}  & \textbf{0.8398$_{(0.0044)}$} & 0.8453$_{(0.0040)}$ & \textbf{0.8777$_{(0.0042)}$}\\
\midrule
\multirow{14}{*}{\textbf{MobileNet-V3}} 
& \textbf{grad\_norm} & 0.0696 & 0.7549$_{(0.0161)}$ & 0.7907$_{(0.0122)}$  & 0.7950$_{(0.0050)}$ & 0.8098$_{(0.0036)}$ & 0.8210$_{(0.0014)}$\\
& \textbf{grasp} & -0.0663 & 0.7440$_{(0.0083)}$ & 0.7894$_{(0.0057)}$  & 0.7918$_{(0.0048)}$ & 0.8116$_{(0.0041)}$ & 0.8183$_{(0.0047)}$\\
& \textbf{jacob\_cov} & -0.0053 & 0.6324$_{(0.0170)}$ & 0.6709$_{(0.0070)}$  & 0.6911$_{(0.0067)}$ & 0.7380$_{(0.0033)}$ & 0.7737$_{(0.0043)}$\\
& \textbf{plain} & 0.0116 & 0.6972$_{(0.0210)}$ & 0.7659$_{(0.0095)}$  & 0.7769$_{(0.0069)}$ & 0.7903$_{(0.0057)}$ & 0.7991$_{(0.0022)}$\\
& \textbf{synflow} & 0.6366 & 0.7355$_{(0.0138)}$ & 0.7930$_{(0.0058)}$  & 0.7935$_{(0.0024)}$ & 0.8127$_{(0.0033)}$ & 0.8172$_{(0.0037)}$\\
& \textbf{synflow\_bn} & 0.6366 & 0.7471$_{(0.0056)}$ & 0.7880$_{(0.0081)}$ & 0.7934$_{(0.0054)}$  & \textbf{0.8149$_{(0.0028)}$} & 0.8186$_{(0.0034)}$ \\
& \textbf{snip} & 0.0725 & 0.7628$_{(0.0076)}$ & 0.7887$_{(0.0125)}$  & 0.7914$_{(0.0052)}$ & 0.8128$_{(0.0059)}$ & \textbf{0.8189$_{(0.0035)}$}\\
& \textbf{fisher} & 0.4199 & 0.6900$_{(0.0116)}$ & 0.7569$_{(0.0038)}$  & 0.7678$_{(0.0036)}$ & 0.7832$_{(0.0031)}$ & 0.7920$_{(0.0017)}$\\
& \textbf{cpu-latency} & 0.6330 & 0.7497$_{(0.0060)}$ & 0.7777$_{(0.0032)}$  & 0.7823$_{(0.0040)}$ & 0.7880$_{(0.0019)}$ & 0.7930$_{(0.0018)}$\\
& \textbf{gpu-latency} & 0.4744 & 0.7519$_{(0.0055)}$ & 0.7720$_{(0.0056)}$ & 0.7765$_{(0.0036)}$ & 0.7908$_{(0.0022)}$ & 0.8071$_{(0.0039)}$\\
& \textbf{parameter-size} & 0.5994 & 0.7549$_{(0.0052)}$ & 0.7933$_{(0.0041)}$  & 0.7941$_{(0.0049)}$ & 0.8070$_{(0.0018)}$ & 0.8088$_{(0.0023)}$\\
\cmidrule(lr){2-8}
& \textbf{vanilla} & - & 0.7373$_{(0.0041)}$ & 0.7852$_{(0.0028)}$  & 0.7832$_{(0.0040)}$ & 0.7944$_{(0.0028)}$ & 0.8023$_{(0.0014)}$\\
& \textbf{ours} & - & \textbf{0.7698$_{(0.0018)}$} & \textbf{0.8034$_{(0.0027)}$}  & \textbf{0.8042$_{(0.0019)}$} & 0.8084$_{(0.0017)}$ & 0.8135$_{(0.0020)}$\\
\midrule
\bottomrule
\end{tabular}}
 \caption{Preliminary experiment results on NDS-ResNet, NDS-ResNeXt-A, and MobileNet-V3. Our proposed dynamic ensemble method utilizes the listed low-fidelity information types for the first-step training. We report Kendall's Tau (average over five runs) achieved by utilizing different types of low-fidelity information with the standard deviation in the subscript. And we make a comparison with the vanilla and dynamic ensemble methods. We encode the architectures with LSTM and train the predictors with the ranking loss. ``Corr.'' represents Kendall's Tau correlation between the low-fidelity information and the actual performance. ``vanilla'' represents directly
training predictor with ground-truth accuracies without low-fidelity information utilization.}
  \label{tab:detailed-nds-mobilenet}
\end{table*}

In addition, we also report the Kendall's Tau correlation between different types of low-fidelity information and the actual performance in Table~\ref{tab:detailed-nds-mobilenet} and Table~\ref{tab:detailed-nasbench}. We find that a high-ranking quality of the low-fidelity information does not indicate its utilization effectiveness. For example, FLOPs (0.5759) is more correlated to the actual performance than snip~\cite{Lee2018snip} (0.4797) on NAS-Bench-201, but utilizing the latter consistently benefits more than utilizing the former with different proportions of training samples. And the parameter size (0.5998) is less correlated to actual performance than the one-shot score (0.6658) on NDS-ResNet, but utilizing the parameter size achieves a higher Kendall's Tau correlation (0.6737) than the one-shot score (0.6296) with 1\% training samples.

Especially, we note that the parameter size and the one-shot score are two of the most beneficial low-fidelity information types on NDS-ResNet.

\begin{table*}[!t]
\centering
\resizebox{\linewidth}{!}{
\begin{tabular}{c|c|c|ccccc}
\toprule
\multirow{2}{*}{\textbf{Search Space}} &\multirow{2}{*}{\textbf{Low-fidelity}} & \multirow{2}{*}{\textbf{Corr.}} &\multicolumn{5}{c}{\textbf{Proportions of training samples}}\\
\cmidrule(lr){4-8} &&& \textbf{1\%}&\textbf{5\%}&\textbf{10\%}&\textbf{50\%}&\textbf{100\%}\\
\midrule
\midrule
\multirow{14}{*}{\textbf{NAS-Bench-201}} 
& \textbf{grad\_norm} & 0.4798 & 0.6601$_{(0.0092)}$ & 0.7559$_{(0.0114)}$  & 0.8088$_{(0.0045)}$ & 0.8817$_{(0.0011)}$ & 0.9005$_{(0.0013)}$\\
& \textbf{grasp} & 0.3227 & 0.5625$_{(0.0045)}$ & 0.6609$_{(0.0085)}$  & 0.7056$_{(0.0076)}$ & 0.8570$_{(0.0036)}$ & 0.8921$_{(0.0013)}$\\
& \textbf{jacob\_cov} & 0.4763 & 0.6705$_{(0.0071)}$ & 0.7655$_{(0.0086)}$  & 0.8031$_{(0.0048)}$ & 0.8769$_{(0.0026)}$ & 0.8980$_{(0.0011)}$\\
& \textbf{one-shot} & 0.6714 & \textbf{0.7844$_{(0.0034)}$} & 0.8269$_{(0.0028)}$  & 0.8496$_{(0.0038)}$ & 0.8900$_{(0.0012)}$ & 0.9044$_{(0.0019)}$\\
& \textbf{plain} & -0.1467 & 0.4882$_{(0.0149)}$ & 0.6213$_{(0.0122)}$  & 0.7092$_{(0.0055)}$ & 0.8469$_{(0.0050)}$ & 0.8845$_{(0.0025)}$\\
& \textbf{relu} & 0.6094 & 0.6829$_{(0.0080)}$ & 0.7641$_{(0.0039)}$  & 0.8001$_{(0.0027)}$ & 0.8781$_{(0.0019)}$ & 0.9013$_{(0.0018)}$\\
& \textbf{relu\_logdet} & 0.6246 & 0.7004$_{(0.0100)}$ & 0.7848$_{(0.0045)}$  & 0.8114$_{(0.0011)}$ & 0.8827$_{(0.0018)}$ & 0.9024$_{(0.0015)}$\\
& \textbf{synflow} & 0.5808 & 0.6525$_{(0.0069)}$ & 0.7523$_{(0.0061)}$  & 0.7958$_{(0.0035)}$ & 0.8796$_{(0.0013)}$ & 0.9005$_{(0.0007)}$\\
& \textbf{snip} & 0.4797 & 0.6594$_{(0.0127)}$ & 0.7545$_{(0.0096)}$  & 0.8058$_{(0.0061)}$ & 0.8800$_{(0.0008)}$ & 0.8995$_{(0.0011)}$\\
& \textbf{fisher} & 0.4152 & 0.6389$_{(0.0124)}$ & 0.7427$_{(0.0066)}$  & 0.8012$_{(0.0038)}$ & 0.8800$_{(0.0019)}$ & 0.8995$_{(0.0013)}$\\
& \textbf{latency} & 0.4860 & 0.5915$_{(0.0119)}$ & 0.6397$_{(0.0026)}$  & 0.6802$_{(0.0048)}$ & 0.8480$_{(0.0051)}$ & 0.8880$_{(0.0020)}$\\
& \textbf{FLOPs} & 0.5759 & 0.5802$_{(0.0045)}$ & 0.6399$_{(0.0049)}$  & 0.7278$_{(0.0101)}$ & 0.8714$_{(0.0028)}$ & 0.8987$_{(0.0010)}$\\
\cmidrule(lr){2-8}
& \textbf{vanilla} & - & 0.5692$_{(0.0087)}$ & 0.6410$_{(0.0018)}$  & 0.7258$_{(0.0053)}$ & 0.8765$_{(0.0010)}$ & 0.9000$_{(0.0008)}$\\
& \textbf{ours} & - & 0.7835$_{(0.0062)}$ & \textbf{0.8538$_{(0.0029)}$}  & \textbf{0.8683$_{(0.0015)}$} & \textbf{0.8992$_{(0.0010)}$} & \textbf{0.9084$_{(0.0010)}$}\\
\midrule
\multirow{12}{*}{\textbf{NAS-Bench-301}} 
& \textbf{grad\_norm} & 0.0378 & 0.5193$_{(0.0143)}$ & 0.6629$_{(0.0091)}$  & 0.7302$_{(0.0022)}$ & 0.7739$_{(0.0033)}$ & 0.7796$_{(0.0069)}$\\
& \textbf{grasp} & 0.4062 & 0.4996$_{(0.0298)}$ & 0.6215$_{(0.0140)}$  & 0.7181$_{(0.0043)}$ & 0.7711$_{(0.0021)}$ & 0.7795$_{(0.0021)}$\\
& \textbf{jacob\_cov} & 0.0958 & 0.4443$_{(0.0317)}$ & 0.6245$_{(0.0106)}$  & 0.7181$_{(0.0040)}$ & 0.7703$_{(0.0023)}$ & 0.7829$_{(0.0016)}$\\
& \textbf{one-shot} & 0.3691 & \textbf{0.5685$_{(0.0105)}$} & 0.6699$_{(0.0053)}$  & 0.7304$_{(0.0047)}$ & 0.7747$_{(0.0031)}$ & 0.7802$_{(0.0015)}$\\
& \textbf{plain} & -0.4670 & 0.2197$_{(0.0701)}$ & 0.6087$_{(0.0249)}$  & 0.7124$_{(0.0040)}$ & 0.7704$_{(0.0044)}$ & 0.7797$_{(0.0014)}$\\
& \textbf{synflow} & 0.1967 & 0.4643$_{(0.0239)}$ & 0.6453$_{(0.0168)}$  & 0.7227$_{(0.0050)}$ & 0.7733$_{(0.0022)}$ & \textbf{0.7844$_{(0.0019)}$}\\
& \textbf{synflow\_bn} & -0.1035 & 0.3683$_{(0.0440)}$ & 0.5830$_{(0.0250)}$  & 0.7025$_{(0.0076)}$ & 0.7662$_{(0.0037)}$ & 0.7788$_{(0.0022)}$\\
& \textbf{snip} & 0.0195 & 0.5219$_{(0.0205)}$ & 0.6707$_{(0.0017)}$  & 0.7329$_{(0.0023)}$ & 0.7741$_{(0.0023)}$ & 0.7822$_{(0.0022)}$\\
& \textbf{fisher} & -0.1847 & 0.4673$_{(0.0592)}$ & 0.6711$_{(0.0111)}$  & 0.7276$_{(0.0033)}$ & 0.7732$_{(0.0025)}$ & 0.7818$_{(0.0018)}$\\
& \textbf{FLOPs} & 0.4619 & 0.5392$_{(0.0189)}$ & \textbf{0.7001$_{(0.0032)}$}  & 0.7412$_{(0.0019)}$ & 0.7672$_{(0.0041)}$ & 0.7880$_{(0.0016)}$\\
\cmidrule(lr){2-8}
& \textbf{vanilla} & - & 0.4160$_{(0.0450)}$ & 0.6752$_{(0.0088)}$  & 0.7354$_{(0.0044)}$ & 0.7693$_{(0.0041)}$ & 0.7883$_{(0.0011)}$\\
& \textbf{ours} & - & 0.5529$_{(0.0135)}$ & 0.6830$_{(0.0038)}$  & \textbf{0.7433$_{(0.0018)}$} & \textbf{0.7752$_{(0.0026)}$} & 0.7842$_{(0.0022)}$\\
\midrule
\bottomrule
\end{tabular}}
 \caption{Preliminary experiment results on NAS-Bench-201 and NAS-Bench-301. Our proposed dynamic ensemble method utilizes the listed low-fidelity information types for the first-step training. We report Kendall's Tau (average over five runs) achieved by utilizing different types of low-fidelity information with the standard deviation in the subscript. And we make a comparison with the vanilla and dynamic ensemble methods. We encode the architectures with GATES and train the predictors with the ranking loss. ``Corr.'' represents Kendall's Tau correlation between the low-fidelity information and the actual performance. ``vanilla'' represents directly training predictor with ground-truth accuracies without low-fidelity information utilization.}
  \label{tab:detailed-nasbench}
\end{table*}

\subsubsection{Comparison with Dynamic Ensemble.} Although inappropriate types of low-fidelity information are utilized in the dynamic ensemble framework, we obverse that they do not damage the prediction ability of the dynamic ensemble predictor. For example, utilizing grasp~\cite{wang2020picking} and jacob\_cov significantly decreases the ranking quality on NDS-ResNet in the preliminary experiment, but our proposed method could still surpass the vanilla method by a large margin.

Moreover, our method can generally perform better than utilizing any single type of low-fidelity information since beneficial knowledge from different types of low-fidelity information can be fused organically. For example, on NAS-Bench-201, our method consistently surpasses only utilizing the one-shot information except with 1\% training samples. In rare cases that the result of our method is not the best, the achieved performance is still no worse than that of utilizing only the most beneficial type of low-fidelity information. For example, with 1\% training samples on NAS-Bench-201, the performance of our method (0.7835) is also basically the same as that of only utilizing the one-shot score (0.7844). 

\begin{table}[!t]
\centering
\begin{tabular}{c|cc}
\toprule
\textbf{Low-fidelity} & \textbf{NDS-ResNet} & \textbf{NDS-ResNeXt-A} \\
\midrule
\midrule
\textbf{grad\_norm} & 0.8825$_{(0.0062)}$ & 0.9450$_{(0.0027)}$ \\
\textbf{grasp} & 0.5050$_{(0.0047)}$ & 0.6099$_{(0.0282)}$ \\
\textbf{jacob\_cov} & 0.6388$_{(0.0170)}$ & 0.5514$_{(0.0158)}$ \\
\textbf{one-shot} & 0.8494$_{(0.0079)}$ & 0.9304$_{(0.0012)}$ \\ 
\textbf{parameter-size} & 0.9627$_{(0.0040)}$ & 0.9695$_{(0.0012)}$ \\
\textbf{plain} & 0.7171$_{(0.0103)}$ & 0.9268$_{(0.0046)}$ \\
\textbf{relu} & 0.9748$_{(0.0005)}$ & 0.9788$_{(0.0002)}$ \\
\textbf{relu\_logdet} & 0.9237$_{(0.0074)}$ & 0.9378$_{(0.0048)}$ \\
\textbf{synflow} & 0.9693$_{(0.0022)}$ & 0.9780$_{(0.0002)}$ \\
\midrule
\bottomrule
\end{tabular}
\caption{Kendall's Tau (average over five runs) between low-fidelity information values and predicted scores after pretraining but before finetuning on NDS-ResNet and NDS-ResNeXt-A, with the standard deviation in the subscript. The architectures are encoded by LSTM.}
\label{tab:kd-between-lf}
\end{table}

\begin{table*}[!t]
\centering
\resizebox{\linewidth}{!}{
\begin{tabular}{c|c|c|ccccc}
\toprule
\multirow{2}{*}{\textbf{Search Space}} &\multirow{2}{*}{\textbf{Low-fidelity}} & \multirow{2}{*}{\textbf{Corr.}} &\multicolumn{5}{c}{\textbf{Proportions of training samples}}\\
\cmidrule(lr){4-8} &&& \textbf{1\%}&\textbf{5\%}&\textbf{10\%}&\textbf{50\%}&\textbf{100\%}\\
\midrule
\midrule
\multirow{11}{*}{\textbf{NDS-ResNet}}
& \textbf{grad\_norm} & 0.2372 & 0.2871$_{(0.0109)}$ & 0.2836$_{(0.0068)}$ & 0.2965$_{(0.0078)}$ & 0.3938$_{(0.0280)}$ & 0.5230$_{(0.0172)}$ \\
& \textbf{grasp} & -0.1142 & -0.1648$_{(0.0168)}$ & -0.1666$_{(0.0131)}$ & -0.1687$_{(0.0157)}$ & -0.0906$_{(0.0339)}$ & 0.0747$_{(0.0456)}$ \\
& \textbf{jacob\_cov} & -0.0724 & -0.0382$_{(0.0175)}$ & -0.0390$_{(0.0164)}$ & -0.0354$_{(0.0161)}$ & 0.0251$_{(0.0152)}$ & 0.1162$_{(0.0192)}$ \\
& \textbf{one-shot} & 0.6658 & 0.6345$_{(0.0110)}$ & 0.6819$_{(0.0105)}$ & 0.7079$_{(0.0113)}$ & 0.7626$_{(0.0151)}$ & 0.7995$_{(0.0126)}$ \\
& \textbf{parameter-size} & 0.5998 & 0.6518$_{(0.0099)}$ & 0.6714$_{(0.0134)}$ & 0.6892$_{(0.0045)}$ & 0.7640$_{(0.0104)}$ & 0.7896$_{(0.0057)}$ \\
& \textbf{plain} & 0.3066 & 0.4180$_{(0.0237)}$ & 0.4005$_{(0.0278)}$ & 0.4090$_{(0.0216)}$ & 0.3332$_{(0.0598)}$ & 0.2647$_{(0.0806)}$ \\
& \textbf{relu} & 0.0733 & 0.0835$_{(0.0041)}$ & 0.0844$_{(0.0028)}$ & 0.0939$_{(0.0066)}$ & 0.1013$_{(0.0446)}$ & 0.1539$_{(0.0674)}$ \\
& \textbf{relu\_logdet} & 0.1817 & 0.2246$_{(0.0139)}$ & 0.2103$_{(0.0190)}$ & 0.2112$_{(0.0216)}$ & 0.2583$_{(0.0202)}$ & 0.3675$_{(0.0203)}$ \\
& \textbf{synflow} & 0.2307 & 0.2913$_{(0.0084)}$ & 0.2793$_{(0.0056)}$ & 0.2969$_{(0.0077)}$ & 0.4288$_{(0.0294)}$ & 0.5605$_{(0.0422)}$ \\
\cmidrule(lr){2-8}
& \textbf{vanilla} & - & 0.2549$_{(0.0087)}$ & 0.4564$_{(0.0108)}$  & 0.5770$_{(0.0094)}$ & 0.7758$_{(0.0078)}$ & 0.8244$_{(0.0110)}$\\
& \textbf{ours} & - & 0.7064$_{(0.0109)}$ & 0.7548$_{(0.0080)}$  & 0.7652$_{(0.0037)}$ & 0.8271$_{(0.0054)}$ & 0.8383$_{(0.0049)}$ \\
\midrule
\multirow{11}{*}{\textbf{NDS-ResNeXt-A}}
& \textbf{grad\_norm} & 0.3190 & 0.3415$_{(0.0067)}$ & 0.3442$_{(0.0070)}$ & 0.3515$_{(0.0095)}$ & 0.3823$_{(0.0137)}$ & 0.4394$_{(0.0191)}$ \\
& \textbf{grasp} & -0.2615 & -0.2978$_{(0.0170)}$ & -0.3018$_{(0.0220)}$ & -0.3102$_{(0.0250)}$ & -0.2717$_{(0.0338)}$ & -0.1675$_{(0.0420)}$ \\
& \textbf{jacob\_cov} & 0.0510 & -0.0181$_{(0.0206)}$ & -0.0346$_{(0.0239)}$ & -0.0520$_{(0.0202)}$ & -0.0033$_{(0.0092)}$ & 0.0465$_{(0.0254)}$ \\
& \textbf{one-shot} & 0.7221 & 0.7375$_{(0.0071)}$ & 0.7571$_{(0.0059)}$ & 0.7812$_{(0.0050)}$ & 0.8140$_{(0.0059)}$ & 0.8244$_{(0.0060)}$ \\
& \textbf{parameter-size} & 0.6683 & 0.6310$_{(0.0086)}$ & 0.6434$_{(0.0156)}$ & 0.6548$_{(0.0131)}$ & 0.7148$_{(0.0092)}$ & 0.7744$_{(0.0102)}$ \\
& \textbf{plain} & 0.2887 & 0.3134$_{(0.0153)}$ & 0.3245$_{(0.0112)}$ & 0.3342$_{(0.0147)}$ & 0.2574$_{(0.1118)}$ & 0.3122$_{(0.0993)}$ \\
& \textbf{relu} & 0.2590 & 0.2156$_{(0.0193)}$ & 0.1811$_{(0.0229)}$ & 0.2103$_{(0.0194)}$ & 0.1868$_{(0.0519)}$ & 0.2195$_{(0.0341)}$ \\
& \textbf{relu\_logdet} & 0.4589 & 0.4278$_{(0.0059)}$ & 0.4319$_{(0.0066)}$ & 0.4364$_{(0.0066)}$ & 0.4755$_{(0.0030)}$ & 0.5496$_{(0.0183)}$ \\
& \textbf{synflow} & 0.6904 & 0.6689$_{(0.0100)}$ & 0.6735$_{(0.0054)}$ & 0.6833$_{(0.0057)}$ & 0.7315$_{(0.0101)}$ & 0.7900$_{(0.0145)}$ \\
\cmidrule(lr){2-8}
& \textbf{vanilla} & - & 0.3568$_{(0.0327)}$ & 0.6243$_{(0.0220)}$  & 0.6671$_{(0.0307)}$ & 0.8224$_{(0.0091)}$ & 0.8701$_{(0.0051)}$\\
& \textbf{ours} & - & 0.7753$_{(0.0010)}$ & 0.8276$_{(0.0024)}$  & 0.8398$_{(0.0044)}$ & 0.8453$_{(0.0040)}$ & 0.8777$_{(0.0042)}$ \\
\midrule
\multirow{14}{*}{\textbf{MobileNet-V3}} 
& \textbf{grad\_norm} & 0.0696 & 0.1849$_{(0.0254)}$ & 0.2587$_{(0.0331)}$ & 0.3226$_{(0.0534)}$ & 0.4725$_{(0.0589)}$ & 0.4475$_{(0.0489)}$ \\
& \textbf{grasp} & -0.0663 & 0.0750$_{(0.0994)}$ & 0.0352$_{(0.0316)}$ & 0.0815$_{(0.0744)}$ & 0.2187$_{(0.0675)}$ & 0.2736$_{(0.0707)}$ \\ 
& \textbf{jacob\_cov} & -0.0053 & 0.0411$_{(0.1305)}$ & 0.3441$_{(0.0377)}$ & 0.3530$_{(0.0437)}$ & 0.4066$_{(0.0383)}$ & 0.4139$_{(0.0512)}$ \\
& \textbf{plain} & 0.0116 & 0.1021$_{(0.1190)}$ & 0.2857$_{(0.0769)}$ & 0.3164$_{(0.0568)}$ & 0.3991$_{(0.0748)}$ & 0.4221$_{(0.0557)}$ \\
& \textbf{synflow} & 0.6366 & 0.6823$_{(0.0049)}$ & 0.7183$_{(0.0074)}$ & 0.7182$_{(0.0051)}$ & 0.7306$_{(0.0093)}$ & 0.7039$_{(0.0344)}$ \\
& \textbf{synflow\_bn} & 0.6366 & 0.6744$_{(0.0078)}$ & 0.7168$_{(0.0109)}$ & 0.7160$_{(0.0072)}$ & 0.7360$_{(0.0116)}$ & 0.7252$_{(0.0181)}$ \\
& \textbf{snip} & 0.0725 & 0.1901$_{(0.0272)}$ & 0.2958$_{(0.0559)}$ & 0.2980$_{(0.0340)}$ & 0.4358$_{(0.0895)}$ & 0.4486$_{(0.0327)}$ \\
& \textbf{fisher} & 0.4199 & 0.4433$_{(0.0252)}$ & 0.5198$_{(0.0184)}$ & 0.5303$_{(0.0203)}$ & 0.5938$_{(0.0174)}$ & 0.6333$_{(0.0202)}$ \\
& \textbf{cpu-latency} & 0.6330 & 0.6548$_{(0.0178)}$ & 0.7048$_{(0.0068)}$ & 0.7045$_{(0.0042)}$ & 0.7083$_{(0.0151)}$ & 0.7053$_{(0.0059)}$ \\
& \textbf{gpu-latency} & 0.4744 & 0.5191$_{(0.0140)}$ & 0.6070$_{(0.0519)}$ & 0.6319$_{(0.0216)}$ & 0.6663$_{(0.0152)}$ & 0.6681$_{(0.0499)}$ \\ 
& \textbf{parameter-size} & 0.5994 & 0.6656$_{(0.0107)}$ & 0.7314$_{(0.0123)}$ & 0.7348$_{(0.0120)}$ & 0.7078$_{(0.0082)}$ & 0.6940$_{(0.0148)}$ \\
\cmidrule(lr){2-8}
& \textbf{vanilla} & - & 0.7373$_{(0.0041)}$ & 0.7852$_{(0.0028)}$  & 0.7832$_{(0.0040)}$ & 0.7944$_{(0.0028)}$ & 0.8023$_{(0.0014)}$\\
& \textbf{ours} & - & 0.7698$_{(0.0018)}$ & 0.8034$_{(0.0027)}$  & 0.8042$_{(0.0019)}$ & 0.8084$_{(0.0017)}$ & 0.8135$_{(0.0020)}$\\
\midrule
\bottomrule
\end{tabular}}
 \caption{The Kendall's Tau correlation (average over five runs) between the weighted score of different low-fidelity experts and the actual performance on NDS-ResNet, NDS-ResNeXt-A and MobileNet-V3. We encode the architectures with LSTM and train the predictors with the ranking loss. ``Corr.'' represents Kendall's Tau correlation between the low-fidelity information and the actual performance. ``vanilla'' represents directly training predictor with ground-truth accuracies without low-fidelity information utilization.}
  \label{tab:detailed-weighted-score}
\end{table*}

\subsubsection{Prediction Ability for Low-fidelity Information.} What has not yet been discussed is whether the predictor can accurately predict low-fidelity information. To answer this question, we report the Kendall's Tau correlation between the predicted score and the low-fidelity information value after pretraining but before finetuning with architectures encoded by LSTM~\cite{luo2018neural} on NDS-ResNet and NDS-ResNeXt-A in Table~\ref{tab:kd-between-lf}. As can be seen, the prediction ability of the predictor for different types of low-fidelity information is different, but all of them are good (KD $\geq$ 0.5050). However, this prediction ability is not directly related to whether utilizing this type of low-fidelity information can lead to improvement. For example, the prediction ability on relu (0.9748) is much higher than that on plain (0.7171) on NDS-ResNet, but utilization of the latter is much more beneficial than the former, as shown in Table~\ref{tab:detailed-nds-mobilenet}. 

\subsection{More Analysis on the Dynamic Ensemble Method}
\subsubsection{Prediction Ability of Each Expert.} The natural question is how is the prediction ability of each expert after training. To answer this question, we show the results on  NDS-ResNet, NDS-ResNeXt-A, and MobileNet-V3 in Table~\ref{tab:detailed-weighted-score}. Specifically, we report the Kendall's Tau correlation between the weighted scores of each low-fidelity expert and the actual performance.

To begin with, we observe that the weighted scores of most low-fidelity experts have a low correlation with the actual performance. For example, the low-fidelity experts of grasp and jacob\_cov have a close to zero or even negative Kendall's Tau coefficient on NDS-ResNet. However, since the standard deviation of their weighted scores is small, as discussed in the main text, these two experts do not significantly degrade the overall performance. In addition, the prediction ability of the entire model is better than any single low-fidelity expert since the entire model is optimized on the overall output and incorporates useful knowledge from different experts.

\begin{table*}[!t]
\centering
\begin{tabular}{c|cccccc}
\toprule
\multirow{2}{*}{\textbf{Search Space}} &\multicolumn{6}{c}{\textbf{Proportions of low-fidelity information training samples}}\\
\cmidrule(lr){2-7} & \textbf{0\%} & \textbf{1\%}&\textbf{5\%}&\textbf{10\%}&\textbf{50\%}&\textbf{100\%}\\
\midrule
\midrule
\textbf{NDS-ResNet} & 0.2549$_{(0.0087)}$ & 0.2518$_{(0.0177)}$ & 0.3952$_{(0.0080)}$ & 0.4864$_{(0.0127)}$ & 0.6783$_{(0.0136)}$ & 0.7064$_{(0.0109)}$ \\
\textbf{NDS-ResNeXt-A} & 0.3568$_{(0.0327)}$ & 0.3982$_{(0.0141)}$ & 0.5984$_{(0.0107)}$ & 0.6588$_{(0.0116)}$ & 0.7688$_{(0.0058)}$ & 0.7753$_{(0.0010)}$ \\
\midrule
\bottomrule
\end{tabular}
 \caption{The Kendall's Tau (average over five runs) of using different proportions of low-fidelity information training samples in the first-step training on NDS-ResNet and NDS-ResNeXt-A. And the standard deviation is in the subscript. The architectures are encoded by LSTM, and the actual performance of the first 1\% architectures by index in the training split is available for the second-step training.}
  \label{tab:diff-pretrain-ratio}
\end{table*}

\begin{figure}[!t]
\center
\includegraphics[width=0.80\linewidth]{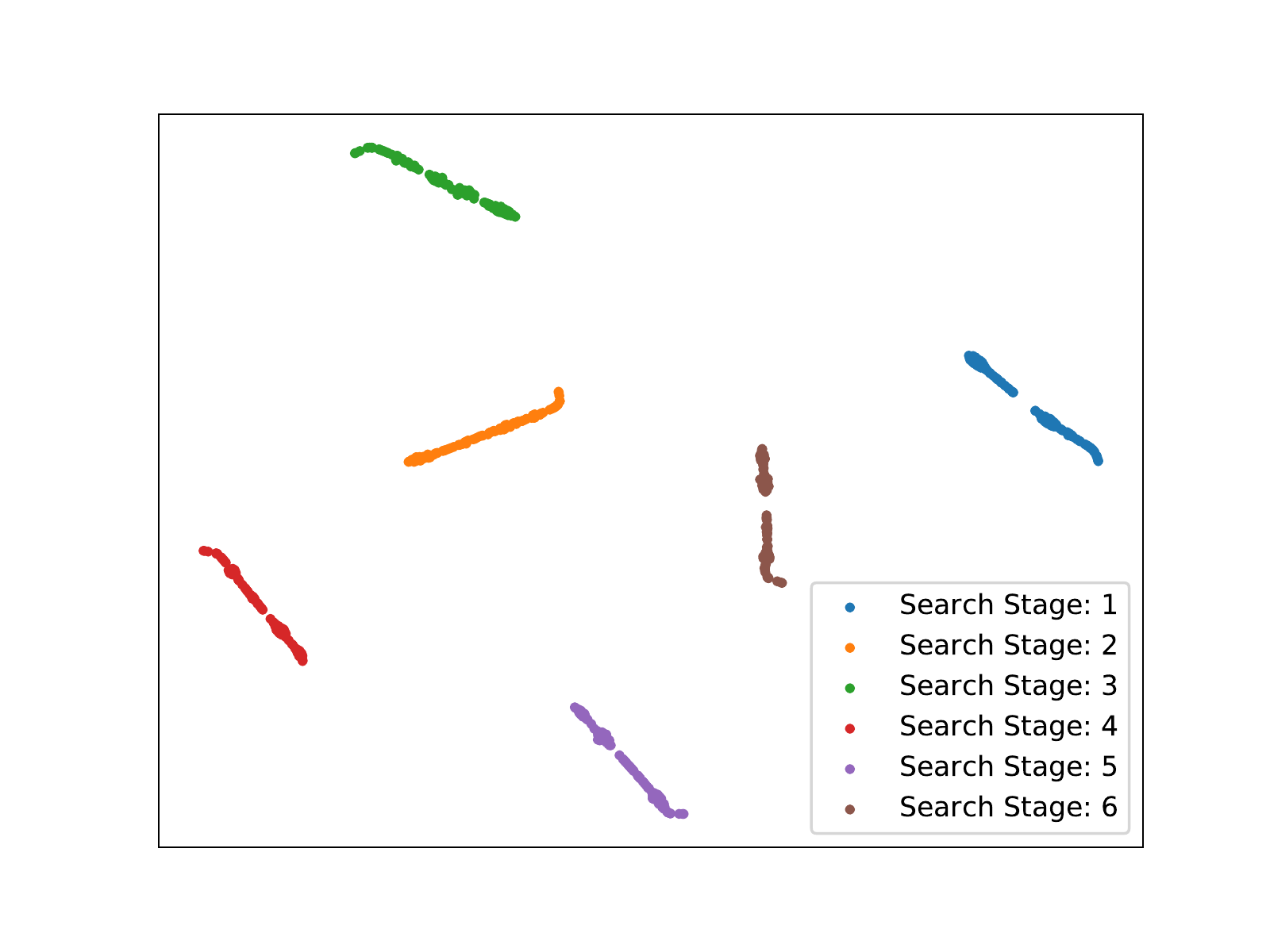}
\caption{T-SNE visualization of weighted coefficients on NAS-Bench-201.}
\label{fig:confidence-tsne}
\end{figure}

\subsubsection{How much low-fidelity information data is required.} Although cheap, evaluation of the low-fidelity information still requires computation cost. Especially for larger search spaces (e.g., DARTS~\cite{liu2018darts}), the computational overhead for some types of low-fidelity information (e.g., one-shot score) can be non-negligible. Therefore, we would like to explore how much low-fidelity information data is required for the first-step training. Specifically, we conduct predictor training on NDS-ResNet and NDS-ResNeXt-A by utilizing the low-fidelity information of different proportions of architectures in the training split, while the actual performance of the first 1\% architectures by index in the training split is available for finetuning.

As shown in Table~\ref{tab:diff-pretrain-ratio}, the final ranking quality generally increases along with the proportion of low-fidelity information data utilized in the first-step training. But even a small amount of data is already beneficial. For example, utilizing 1\% low-fidelity information data improves the Kendall's Tau from 0.3568 to 0.3982 on NDS-ResNeXt-A. This indicates that in practice, sampling a small number of architectures and acquiring their low-fidelity information could also be beneficial.

\begin{figure}[!t]
\center
\includegraphics[width=0.95\linewidth]{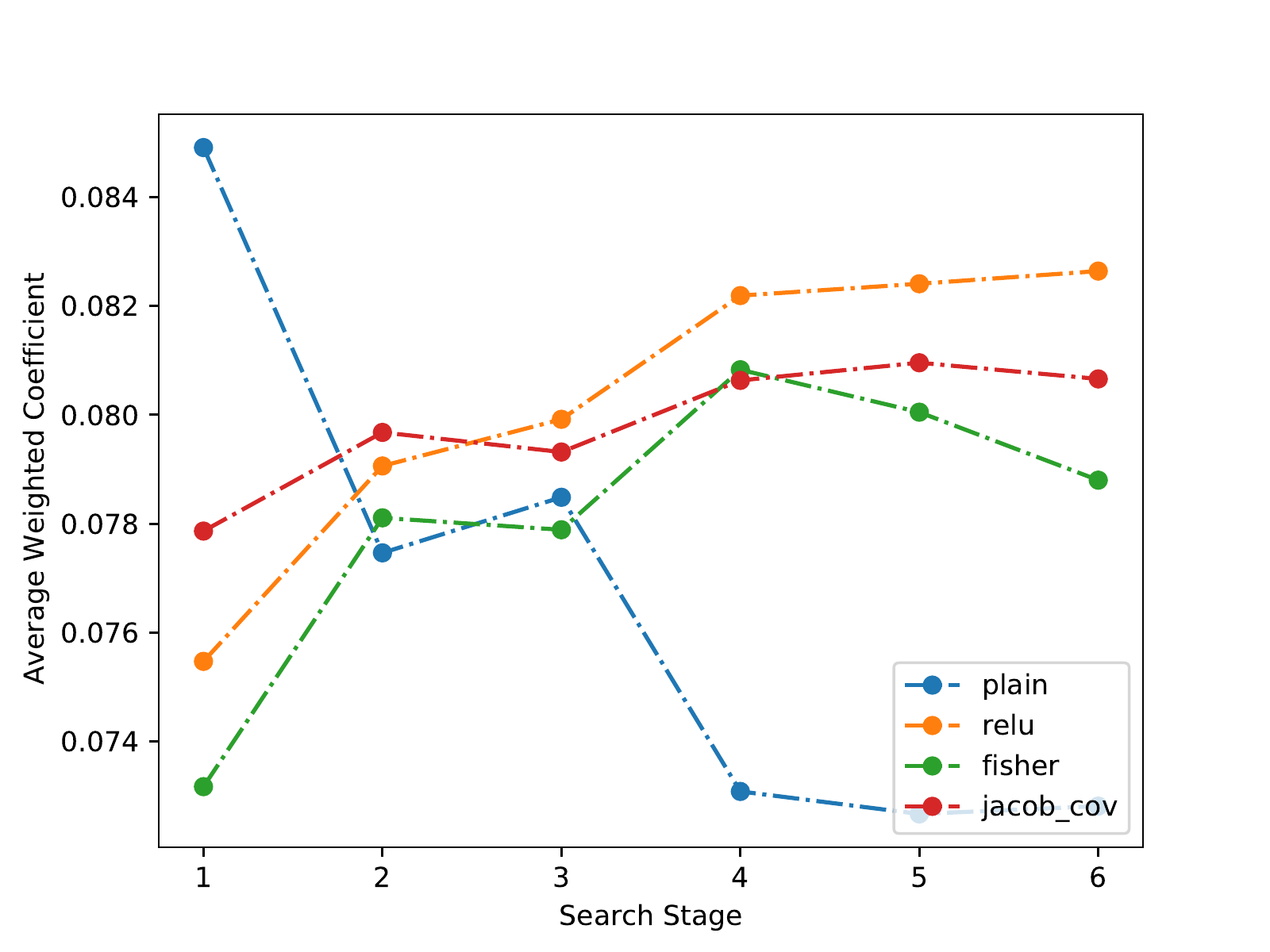}
\caption{Average weighted coefficients of different low-fidelity experts at different search stages.}
\label{fig:confidence-change}
\end{figure}

\begin{table}[!t]
\centering
\resizebox{\linewidth}{!}{
\begin{tabular}{c|c|ccc}
\toprule
\multirow{2}{*}{\textbf{Search Space}}
&\multirow{2}{*}{\textbf{Ensemble}} &\multicolumn{3}{c}{\textbf{Proportions of training samples}}\\
\cmidrule(lr){3-5} && \textbf{1\%}&\textbf{5\%}&\textbf{10\%}\\
\midrule
\midrule
\multirow{2}{*}{\textbf{NDS-ResNet}}
& \textbf{vanilla} & 0.6110$_{(0.0230)}$ & 0.6621$_{(0.0149)}$ & 0.6530$_{(0.0086)}$ \\
& \textbf{dynamic} & \textbf{0.7064$_{(0.0109)}$} & \textbf{0.7548$_{(0.0080)}$} & \textbf{0.7652$_{(0.0037)}$} \\
\midrule
\multirow{2}{*}{\textbf{NDS-ResNeXt-A}}
& \textbf{vanilla} & 0.7510$_{(0.0096)}$ & 0.8027$_{(0.0055)}$  & 0.8184$_{(0.0031)}$ \\
& \textbf{dynamic} & \textbf{0.7753$_{(0.0010)}$} & \textbf{0.8276$_{(0.0024)}$}  & \textbf{0.8398$_{(0.0044)}$}\\
\midrule
\midrule
\bottomrule
\end{tabular}}
 \caption{Comparison between the vanilla ensemble and the dynamic ensemble method on NDS-ResNet and NDS-ResNeXt-A. We report the Kendall's Tau (average over five runs) between the predicted score and actual performance, and the standard deviation is in the subscript. The architectures are encoded by LSTM.}
  \label{tab:comparison-vanilla-ensemble}
\end{table}

\subsubsection{Comparison with Vanilla Ensemble.} To further verify the effectiveness of the dynamic ensemble, we remove the gating network and assign the same weight coefficients to different low-fidelity experts in the second-step training. We simply refer to this method as vanilla ensemble. We conduct experiments on NDS-ResNet and NDS-ResNeXt-A to make the comparison. As shown in Table~\ref{tab:comparison-vanilla-ensemble}, compared with the dynamic ensemble, the vanilla ensemble method significantly reduces the experimental results. For example, the Kendall's Tau correlation achieved by the vanilla ensemble method with 1\% training samples on NDS-ResNet is only 0.6110, much lower than that of the dynamic ensemble method (0.7064).

\begin{table*}[!t]
\centering
\resizebox{\linewidth}{!}{
\begin{tabular}{c|c|ccccc}
\toprule
\multirow{2}{*}{\textbf{Search Space}}
&\multirow{2}{*}{\textbf{Manner}} &\multicolumn{5}{c}{\textbf{Proportions of training samples}}\\
\cmidrule(lr){3-7} && \textbf{1\%}&\textbf{5\%}&\textbf{10\%}&\textbf{50\%}&\textbf{100\%}\\
\midrule
\midrule
\multirow{3}{*}{\textbf{NAS-Bench-201}}
& \textbf{random forest} & 0.4151$_{(0.0066)}$ & 0.5886$_{(0.0042)}$ & 0.6310$_{(0.0029)}$ & 0.7038$_{(0.0041)}$ & 0.7023$_{(0.0055)}$\\
& \textbf{ours (LSTM)} & 0.7835$_{(0.0062)}$ & 0.8538$_{(0.0029)}$  &0.8683$_{(0.0015)}$ & 0.8992$_{(0.0010)}$ & 0.9084$_{(0.0010)}$\\
& \textbf{ours (GATES)} & \textbf{0.8244$_{(0.0081)}$} & \textbf{0.8948$_{(0.0021)}$}  & \textbf{0.9075$_{(0.0015)}$} & \textbf{0.9216$_{(0.0019)}$} & \textbf{0.9250$_{(0.0020)}$}\\
\midrule
\multirow{3}{*}{\textbf{NAS-Bench-301}}
& \textbf{random forest} & 0.3549$_{(0.0052)}$ & 0.5020$_{(0.0088)}$ & 0.5220$_{(0.0045)}$ & 0.6199$_{(0.0035)}$ & 0.6418$_{(0.0015)}$\\
& \textbf{ours (LSTM)} & 0.4805$_{(0.0083)}$ & 0.6405$_{(0.0035)}$  & 0.7075$_{(0.0022)}$ & 0.7544$_{(0.0028)}$ & 0.7751$_{(0.0011)}$\\
& \textbf{ours (GATES)} & \textbf{0.5529$_{(0.0135)}$} & \textbf{0.6830$_{(0.0038)}$}  & \textbf{0.7433$_{(0.0018)}$} & \textbf{0.7752$_{(0.0026)}$} & \textbf{0.7842$_{(0.0022)}$}\\
\midrule

\multirow{2}{*}{\textbf{NDS-ResNet}}
& \textbf{random forest} & 0.3249$_{(0.0167)}$ & 0.5653$_{(0.0056)}$  & 0.6494$_{(0.0025)}$ & 0.7050$_{(0.0041)}$ & 0.7473$_{(0.0029)}$\\
& \textbf{ours (LSTM)} & \textbf{0.7064$_{(0.0109)}$} & \textbf{0.7548$_{(0.0080)}$}  & \textbf{0.7652$_{(0.0037)}$} & \textbf{0.8271$_{(0.0054)}$} & \textbf{0.8383$_{(0.0049)}$}\\
\midrule

\multirow{2}{*}{\textbf{NDS-ResNeXt-A}}
& \textbf{random forest} & 0.2967$_{(0.0198)}$ & 0.5583$_{(0.0038)}$  & 0.6107$_{(0.0037)}$ & 0.7244$_{(0.0036)}$ & 0.7332$_{(0.0025)}$\\
& \textbf{ours (LSTM)} & \textbf{0.7753$_{(0.0010)}$} & \textbf{0.8276$_{(0.0024)}$}  & \textbf{0.8398$_{(0.0044)}$} & \textbf{0.8453$_{(0.0040)}$} & \textbf{0.8777$_{(0.0042)}$}\\
\midrule

\multirow{2}{*}{\textbf{MobileNet-V3}} 
& \textbf{random forest} & 0.6411$_{(0.0057)}$ & 0.7018$_{(0.0019)}$  & 0.7179$_{(0.0002)}$ & 0.7451$_{(0.0002)}$ & 0.7529$_{(0.0002)}$\\
& \textbf{ours (LSTM)} & \textbf{0.7698$_{(0.0018)}$} & \textbf{0.8034$_{(0.0027)}$}  & \textbf{0.8042$_{(0.0019)}$} & \textbf{0.8084$_{(0.0017)}$} & \textbf{0.8135$_{(0.0020)}$}\\
\midrule
\bottomrule
\end{tabular}}
 \caption{The Kendall's Tau (average over five runs) of using different methods and encoders on NAS-Bench-201, NAS-Bench-301, NDS-ResNet, NDS-ResNeXt-A and MobileNet-V3. And the standard deviation is in the subscript.}
  \label{tab:random-forest}
\end{table*}

\begin{table*}[!t]
\centering
\begin{tabular}{c|ccc|c}
\toprule
\textbf{Constraint}
&\textbf{Vanilla} &\textbf{Ours} &\textbf{Random Sample} &\textbf{Optimal}\\
\midrule
\midrule
\textbf{$\leq$ 50M} & 0.9292$_{(0.0017)}$ & \textbf{0.9331$_{(0.0008)}$} & 0.9160$_{(0.0089)}$ & 0.9348 \\
\textbf{$\leq$ 75M} & 0.9334$_{(0.0045)}$ & \textbf{0.9357$_{(0.0000)}$} & 0.9225$_{(0.0045)}$ & 0.9375 \\
\textbf{$\leq$ 100M} & 0.9387$_{(0.0011)}$ & \textbf{0.9427$_{(0.0007)}$} & 0.9295$_{(0.0080)}$ & 0.9431 \\
\bottomrule
\end{tabular}
\caption{Discovered architecture accuracies under various FLOPs constraints on NAS-Bench-201. We report the average values over ten runs with different seeds, and the standard deviation is in the subscript.}
\label{tab:nb201-search-constraints}
\end{table*}

\subsubsection{Inspection into Different Search Stages.} We visualize the updates of weighted coefficients through search stages in Figure~\ref{fig:confidence-tsne}. Specifically, we collect 100 architectures from the validation split of NAS-Bench-201. We use t-SNE to map their weighted coefficients to a 2-dim space and visualize them. We can see that the dynamic ensemble method keeps refining the weighted coefficients across search stages. 

Interestingly, we observe that the weighted coefficients of some low-fidelity experts have a consistent trend between different search stages. We show how the average of the weighted coefficients of some low-fidelity experts changes over the search phase in Figure~\ref{fig:confidence-change}. For example, the weighted coefficients of the expert ``plain'' tend to decrease while those of the expert ``relu'' tend to increase through the search process.

\subsection{Comparison with Random Forest}
We compare the prediction ability of our proposed dynamic ensemble predictor with the random forest~\cite{breiman2001random} algorithm. In particular, we use the RandomForestRegressor from scikit-learn~\cite{scikit-learn} for our implementation. We set the number of trees in the random forest (n\_estimatores) to 100, use MSE to measure the quality of a split, and set "max\_features" to 0.5. Other hyperparameters use default values. Each architecture is encoded as a sequence in the simplest way and fed into the random forest to get the prediction score.

As the results shown in Table~\ref{tab:random-forest}, our method consistently outperforms random forest by a large margin. For example, on NDS-ResNet and NDS-ResNeXt-A, our method achieves 0.7064 and 0.7753 Kendall's Tau with 1\% training samples, respectively, much better than random forest (0.3249, 0.2967).

\subsection{Architecture Search under Budget Constrains}
Our proposed method can be easily extended to discover high-performance architectures under resource constraints by only sampling architectures that meet the budget limitation. To demonstrate the effectiveness of our method in constrained NAS, we conduct experiments on NAS-Bench-201 and the DARTS search space, respectively. Without loss of generality, we take the number of floating-point operations (FLOPs) as the resource indicator. 

\subsubsection{NAS-Bench-201.} We search for architectures with FLOPs smaller than 50M, 75M, and 100M, respectively. Specifically, we finetune the entire predictor on 1\% architectures in the training split after the first training step. Then, we traverse all the architectures satisfying the FLOPs constraint in the search space. The best test accuracy among the top-10 architectures selected by the predictor is reported. We compare our method with the vanilla predictor-based method and random sampling. As shown in Table~\ref{tab:nb201-search-constraints}, our discovered architectures surpass those discovered by other search strategies under various constraints.

\subsubsection{DARTS Search Space.} We compare our methods with SNAS~\cite{xie2018SNAS} and GDAS~\cite{dong2019GDAS} on CIFAR-10 by searching for architectures with FLOPS fewer than 474M or 545M, respectively. Our discovered two architectures achieve 97.31\% (473.5M) and 97.40\% (544M), respectively, surpassing SNAS (97.15\%, 474M) and GDAS (97.07\%, 545M) for a large margin with fewer FLOPs.

\begin{figure}[!t]
\center
\includegraphics[width=\linewidth]{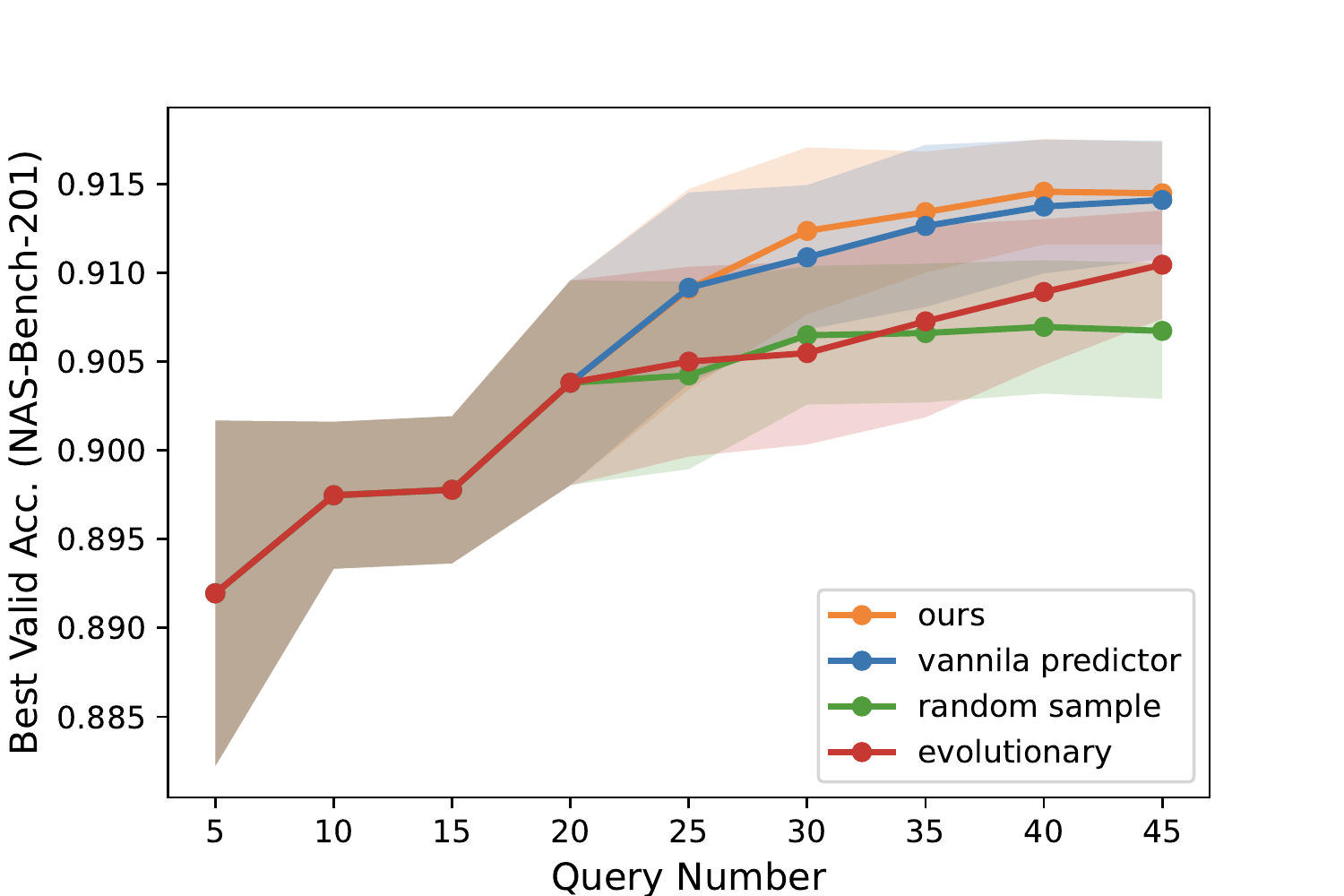}
\caption{Comparison with other search strategies on NAS-Bench-201. We run each experiment 10 times with different seeds and report the best validation accuracy of all sampled architectures.}
\label{fig:search_nb201_val}
\end{figure}

\begin{figure*}[!h]
\center
\includegraphics[width=\linewidth]{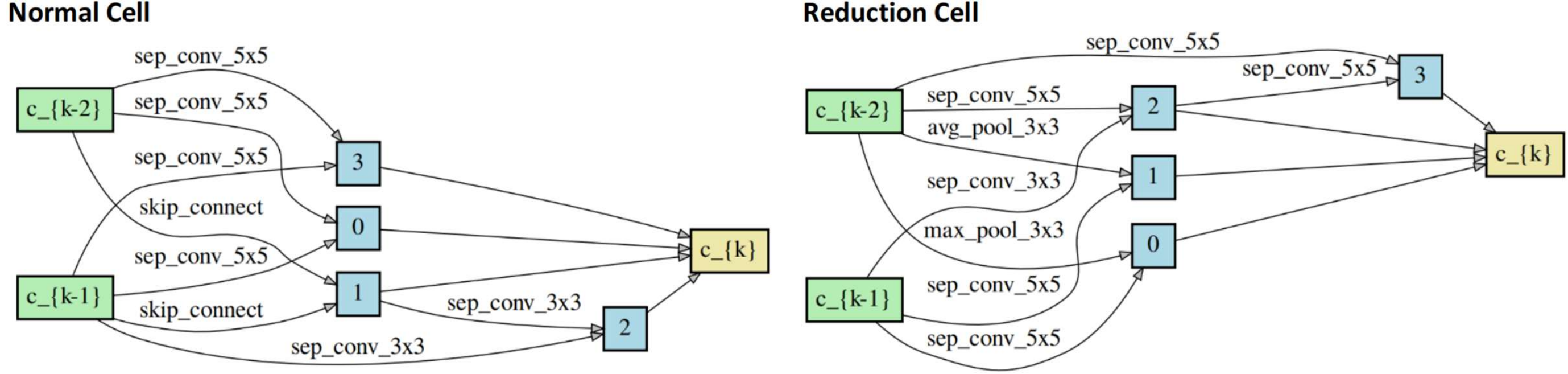}
\caption{Our discovered architecture in the DARTS search space.}
\label{fig:arch}
\end{figure*}

\subsection{Validation Accuracy on NAS-Bench-201}
As the gap exists between test accuracies and validation accuracies on NAS-Bench-201, we also show the highest validation accuracies of all sampled architectures during the search process in Fig.~\ref{fig:search_nb201_val}. Our search flow achieves the highest any-time validation accuracy, demonstrating the proposed method's effectiveness.

\subsection{Discovered Arch. in the DARTS Search Space} We illustrate the discovered architecture in the DARTS search space in Figure~\ref{fig:arch}.

\clearpage

\section{Detailed Experimental Settings}

\subsection{Search Space}
We conduct experiments on the following four NAS benchmarks for a thorough evaluation.

\subsubsection{NAS-Bench-201~\cite{dong2020bench}.} A tabular benchmark that contains the complete training information of 15625 architectures in a cell-based NAS search space. In our experiments, we use 50\% (7813) architectures in the search space as the training split and the other 50\% (7812) as the validation split.

\subsubsection{NAS-Bench-301~\cite{siems2020bench}.} A surrogate benchmark that predicts the performances of $10^{18}$ architectures in DARTS\cite{liu2018darts} search space, with the stand-alone performances of 60k landmark architectures. To acquire a more accurate evaluation, we randomly sample 5896 architectures from the landmark set to construct the training split and use the other (51072) architectures as the validation split. 

\subsubsection{NDS ResNet / ResNeXt-A~\cite{radosavovic2019nds}.} A tabular benchmark that provides training accuracy of partial architectures in the ResNet search space and the ResNeXt-A search space. The ResNet search space enables search for non-topological decisions, including depth and width of architectures, while the ResNeXt-A search space additionally contains the bottleneck width ratio and the number of groups. In our experiments, both the training split and the validation split contain 2500 randomly sampled architectures in each search space.

\subsubsection{MobileNetV3~\cite{cai2019once}.} A benchmark that provides accuracy of architectures on ImageNet in the MobileNetV3~\cite{howard2019searching} design space. To avoid the prohibitively high computational cost of training each architecture from scratch, the once-for-all~\cite{cai2019once} technique is adopted by evaluating architectures with their corresponding weights in a pre-trained supernet. In our experiments, both the training split and the validation split contain 10000 randomly sampled architectures.

\subsection{Low-fidelity Estimation \& Acquisition}
The low-fidelity information types utilized in our paper are listed as follows. Specifically, the utilized types of low-fidelity information for each search space are listed in Table~\ref{tab:detailed-nds-mobilenet} and Table~\ref{tab:detailed-nasbench}.

\subsubsection{One-shot Estimation.} We acquire the one-shot estimation following Ning \emph{etal.}~\cite{ning2020surgery} All the training and evaluation are conducted on CIFAR-10. The original training dataset of CIFAR-10 is divided into two parts: training split (40000 images) and validation split (10000 images). The supernets are trained on the training split, and architecture rewards are evaluated on the validation split. We train all the supernets using an SGD optimizer with a momentum of 0.9 and a weight decay of 0.0005 for 1000 epochs. The learning rate is set to 0.05 initially and decayed by 0.5 each time the supernet’s average training loss stops to decrease for 30 epochs. The batch size is set to 512 / 256 / 64 on NAS-Bench-201, NAS-Bench-301 and NDS-ResNet / ResNeXt-A, respectively. The $l_2$ norm of the gradient is clipped to 5.0, and a drop rate of 0.1 is applied. 

\subsubsection{Zero-shot Estimation.} Zero-shot estimations studied in this paper include grad\_norm, grasp, jacob\_cov, plain, relu, relu\_logdet~\cite{mellor2021neural}, synflow, synflow\_bn~\cite{tanaka2020pruning}, fisher and snip. We evaluate these zero-shot estimations following Ning \emph{etal.}~\cite{ning2020surgery} Specifically, we use a batch size of 128 on NAS-Bench-201 / NAS-Bench-301 / NDS-ResNet / MobileNet-V3 and a batch size of 64 on the NDS-ResNeXt-A search space. We evaluate the ZSEs with five batches in total and calculate the average score of 5 validation batches as the result.

\subsubsection{Complexity Estimation.} Complexity estimations studied in this paper include the parameter size,  the number of
floating-point operations (FLOPs) and latency. Specifically, we study the Intel Xeon CPU latency (cpu-latency) and NVIDIA 1080Ti latency (gpu-latency) on MobileNet-V3 following once-for-all~\cite{cai2019once}.

\subsection{Predictor Construction}
A predictor first encodes the architecture into a continuous vector and then feeds the vector into an MLP to get the prediction score. We give out the detailed predictor construction and training settings on different search spaces.

\subsubsection{NAS-Bench-201.} We experiment with two types of architecture encoders. 1) \textbf{LSTM}~\cite{luo2018neural}: The embedding size and hidden size of the 1-layer LSTM are set to 100, and the final hidden state is used as the embedding of the cell architecture. The encoding vector is fed into a 3-layer MLP with 200 hidden units to get the final score; 2) \textbf{GATES}~\cite{ning2020generic}: The operation embedding dimension, node embedding dimension, and hidden dimension are 48, 48, and 96, respectively. Five 128-dimension GCN layers are used. The encoding vector is fed into a 5-layer MLP with 300 hidden units to get the final score.

\subsubsection{NAS-Bench-301.} We apply LSTM as the architecture encoder. The embedding size and hidden size of the 1-layer LSTM are set to 48 and 128, respectively. The final hidden state is used as the cell architecture embedding. The encoding vector is fed into a 200-dimension fully-connected layer to get the final score. 

\subsubsection{NDS-ResNet, NDS-ResNeXt and MobileNetV3.} We apply LSTM as the architecture encoder. The embedding size and hidden size of the 1-layer LSTM are set to 100, and the final hidden state is used as the embedding of the cell architecture. The encoding vector is fed into a 3-layer MLP with 200 hidden units to get the final score. 

\subsection{Predictor Training}
We first train different low-fidelity experts for 200 epochs and then finetune the dynamic ensemble performance predictor on the actual performance data for 200 epochs. In contrast, we directly train the vanilla predictor on the actual performance data for 200 epochs. For the preliminary experiment in the introduction, we pretrain and then finetune the predictor both for 200 epochs. Specially, we re-initialize the optimizer before finetuning. 

We apply an Adam optimizer with a learning rate of 0.001. A dropout of 0.1 is used to train the architecture encoder. The batch sizes used for NAS-Bench-201, NAS-Bench-301, NDS, and MobileNetV3 search spaces are 512, 128, 128, and 512, respectively. 

\subsection{Evaluation on the DARTS Search Space}
We show our discovered architecture in the DARTS search space in Figure~\ref{fig:arch}. For the final comparison of different architectures, we train our discovered architectures on CIFAR-10, CIFAR-100, and ImageNet, respectively.

On CIFAR-10 and CIFAR-100, the initial channel number of the discovered architecture is augmented to 36. The standard data pre-processing and augmentation techniques as in previous works~\cite{liu2018darts} are applied. The architecture is trained for 600 epochs, using a batch size of 128, weight decay of 3e-4, and an SGD optimizer with a momentum of 0.9. The initial learning rate is 0.05, decayed down to 0 following a cosine schedule. The $l_2$ norm of the gradient is clipped to 5.0, and a drop rate of 0.1 is applied. Additional enhancements as in previous works~\cite{liu2018darts} include cutout~\cite{2017Improved}, path dropout of probability 0.2, and auxiliary towers with a weight of 0.4.

On ImageNet, the initial channel number of the discovered architecture is augmented to 48. The data augmentation techniques are set to be the same as those in~\cite{liu2018darts}. The discovered architecture is trained for 250 epochs, using a batch size of 256, weight decay of 3e-5, and an SGD optimizer with a momentum of 0.9. The initial learning rate is 0.1, decayed by a factor of 0.97 after each epoch. The $l_2$ norm of the gradient is clipped to 5.0. Additional enhancements include path dropout of probability 0.2, auxiliary towers with a weight of 0.4, and label smoothing regularization~\cite{2016Rethinking} with a weight of 0.1.

\end{appendices}

\end{document}